%% file: main.tex
\documentclass[journal]{IEEEtran}
\usepackage{amsmath,amsfonts}
\usepackage{algorithmic}
\usepackage{algorithm}
\usepackage{array}
\usepackage[caption=false,font=normalsize,labelfont=sf,textfont=sf]{subfig}
\usepackage{textcomp}
\usepackage{stfloats}
\usepackage{url}
\usepackage{verbatim}
\usepackage{graphicx}
\usepackage{cite}

\usepackage{multirow}
\usepackage[table]{xcolor}
\usepackage{amsmath}
\usepackage{amsthm}
\usepackage{booktabs}
\usepackage{hyperref}
\usepackage[T1]{fontenc}
\usepackage[utf8]{inputenc}
\usepackage{circledsteps}
\usepackage{pifont}

\begin{document}

\title{Adversarial Example Soups: Improving Transferability and Stealthiness for Free}

\author{Bo Yang$^*$,
        Hengwei Zhang$^*$,
        Jindong Wang,
        Yulong Yang,
        Chenhao Lin,
        Chao Shen,
        Zhengyu Zhao
\thanks{Bo Yang, Hengwei Zhang, and Jindong Wang are with the State Key Laboratory of Mathematical Engineering and Advanced Computing and Henan Key Laboratory of Information Security, Information Engineering University, Zhengzhou, China.}
\thanks{Yulong Yang, Chenhao Lin, Chao Shen, and Zhengyu Zhao are with the School of Cyber Science and Engineering, Xi'an Jiaotong University, Xi'an, China.}
\thanks{$^*$The first two authors contribute equally.}

\thanks{Corresponding authors: Hengwei Zhang (wlby\_zzmy\_henan@163.com) and Zhengyu Zhao (zhengyu.zhao@xjtu.edu.cn)}
}

\maketitle

\begin{abstract}

Transferable adversarial examples cause practical security risks since they can mislead a target model without knowing its internal knowledge.
A conventional recipe for maximizing transferability is to keep only the optimal adversarial example from all those obtained in the optimization pipeline.
In this paper, for the first time, we revisit this convention and demonstrate that those discarded, sub-optimal adversarial examples can be reused to boost transferability.
Specifically, we propose ``Adversarial Example Soups'' (AES), with AES-tune for averaging discarded adversarial examples in hyperparameter tuning and AES-rand for stability testing. In addition, our AES is inspired by ``model soups'', which averages weights of multiple fine-tuned models for improved accuracy without increasing inference time.
Extensive experiments validate the global effectiveness of our AES, boosting 10 state-of-the-art transfer attacks and their combinations by up to 13\% against 10 diverse (defensive) target models. We also show the possibility of generalizing AES to other types, \textit{e.g.}, directly averaging multiple in-the-wild adversarial examples that yield comparable success. A promising byproduct of AES is the improved stealthiness of adversarial examples since the perturbation variances are naturally reduced. The code is available at \url{https://github.com/yangbo93/Adversarial-Example-Soups}.

\end{abstract}


\section{Introduction}
\label{sec:intro}

In recent years, Deep Neural Networks (DNNs) have achieved great success in various domains, such as image classification~\cite{b1,b2}, face recognition~\cite{b3,b4}, object detection~\cite{b5,chen2022transzero++,huang2023anti}, and autonomous driving~\cite{zhu2024sora,muhammad2020deep}.
However, DNNs are known to be vulnerable to adversarial examples~\cite{b7,b8,b9,b10,b11}, which are crafted by adding imperceptible perturbations into clean images.
Adversarial examples can cause severe threats in black-box security-sensitive applications, such as face recognition systems~\cite{lin2022real} and autonomous driving cars~\cite{zhang2018camou}, due to their transferability, \textit{\textit{i.e.},} the adversarial examples generated on the surrogate model can be directly used to mislead unknown target models~\cite{b13,zhao2023adversarial,b14,b15,b16}.

\begin{figure}[!t]
  \centering
    \includegraphics[width=0.95\linewidth]{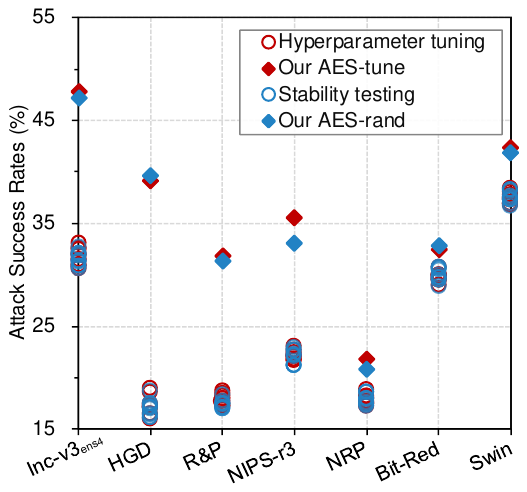}
    \caption{Our Adversarial Example Soups (AES) attack consistently improves the transferability from Inc-v3 to diverse (defensive) target models on ImageNet. AES-tune (AES-rand) averages 10 sessions of adversarial examples from hyperparameter tuning (stability testing). Here we report the results for the well-known DIM~\cite{b36} attack and other attacks also yield similar patterns.}
    \label{fig1}
\end{figure}

Transferable adversarial examples have been extensively studied~\cite{b17,b18,yang2024quantization,b19,b20,zhao2021success,liu2019s}, with new attack methods continuously proposed based on diverse ideas, such as gradient stabilization~\cite{b21,b22}, input transformation~\cite{b23,b24}, and feature disruption~\cite{b25,b26}.
In general, generating a (transferable) adversarial example is a gradient-based optimization problem.
Therefore, when a new transferable attack is proposed, existing work follows the conventional optimization recipe to conduct hyperparameter tuning and stability testing. 
Specifically, hyperparameter tuning is conducted to select the hyperparameter(s) that ensure optimal transferability, and stability testing is conducted to study the sensitivity of the proposed attack to potential randomness in the whole attack pipeline.
We examine 30 representative transfer studies and notice that hyperparameter tuning is conducted in all of them, and stability testing is also often conducted, \textit{e.g.}, in~\cite{wang2020unified,b35,zhang2024does}.

Such a conventional optimization recipe normally consumes substantial resources of computation and time.
However, existing attacks keep only the final optimal adversarial example but discard all the rest contained through the process.
In this paper, for the first time, we revisit this conventional recipe and demonstrate that it is possible to \textit{``turn trash into treasure''}, \textit{i.e.}, leveraging those discarded, sub-optimal adversarial examples to boost transferability for free.

To this end, we propose ``\textbf{Adversarial Example Soups}'' (AES), which reuse multiple discarded adversarial examples by simply averaging them, including the optimal one.
Specifically, AES-tune reuses the discarded samples in hyperparameter tuning, and AES-rand targets stability testing.
As shown in Figure~\ref{fig1}, both AES-tune and AES-rand improve their corresponding baselines, i.e., hyperparameter tuning and stability testing, by a large margin.

For the averaging operation, we explore three common strategies, \textit{i.e.}, uniform, weighted, and greedy.
Extensive experiments involving 10 state-of-the-art transferable attacks and 10 diverse (defensive) target models validate the global effectiveness of our AES.
We also investigate a practically promising scenario where multiple in-the-wild adversarial examples with a similar level of success from arbitrary attacks can also be averaged to boost their transferability.
In addition to improving attack transferability, AES improves attack stealthiness because the perturbation variances are naturally reduced through the averaging operation.

AES is inspired by ``model soups''~\cite{b27}, which can improve the accuracy and robustness of a model by simply averaging the weights of multiple models fine-tuned with different hyperparameter configurations.
This idea can work in the context of transferability for the following reasons.
Overall, it is well known that transferring an adversarial example from a known (surrogate) model to an unknown (target) model is conceptually similar to generalizing a model from known (training) examples to unknown (testing) examples~\cite{b21,b36}.
In this case, optimizing adversarial perturbations is analogous to optimizing model weights.
In particular, the underlying assumption of ``model soups'' is that the weights of multiple fine-tuned models often lie in a single error basin~\cite{b27,b28,b29}, which also holds in our context of adversarial attacks (see experimental evidence in Section~\ref{sec:motivation}).

In sum, we make the following main contributions:

\begin{itemize}

\item We, for the first time, revisit the conventional recipe for optimizing adversarial transferability, which only picks the optimal adversarial example but discards the rest (sub-optimal) ones during hyperparameter tuning and stability testing.
Instead, we \textit{``turn trash into treasure''}, \textit{i.e.}, leveraging those discarded adversarial examples to boost transferability for free.

\item We propose ``Adversarial Example Soups'' (AES), a new approach to reusing multiple discarded adversarial examples by averaging them and the optimal one.
We demonstrate the global effectiveness of AES in improving 10 state-of-the-art transferable attacks and their combinations on 10 diverse (defensive) target models.

\item We further demonstrate the effectiveness of AES in improving attack stealthiness, and we discuss other potential types of AES, \textit{e.g.}, improving transferability by averaging multiple in-the-wild adversarial examples with comparable success rates generated by arbitrary attacks.

\end{itemize}

\section{Related Work}
\label{sec:related}

\subsection{Adversarial Attacks}
\label{sec:adv-attack}

Various methods have been developed to enhance the transferability of adversarial examples, and they can be divided into the following six types.

\noindent\textbf{Gradient stability attacks} improve the adversarial attack from the perspective of gradient optimization, thereby enhancing the convergence effect of the attack process. Typical methods include incorporating momentum~\cite{b21}, nesterov accelerated gradient~\cite{b22}, variance tuning~\cite{b34}, and penalizing gradient norm~\cite{b35} into the attack process.

\noindent\textbf{Input transformation attacks} alleviate the overfitting of adversarial attacks from the perspective of data augmentation, thereby improving adversarial transferability. Typical methods focus on diverse inputs~\cite{b36}, scale invariance~\cite{b22}, images mixing~\cite{b37}, and frequency-domain transformations~\cite{b38}.

\noindent\textbf{Feature disruption attacks} enhance the attack process by manipulating the features that dominate classification. Typical methods include Feature Disruptive Attack~\cite{b39}, Feature Importance-aware Attack~\cite{b25}, and Neuron Attribution-Based Attack~\cite{b26}.

\noindent\textbf{Model modification attacks} optimize transfer-based attacks by adjusting the surrogate model. Typical methods construct ghost networks~\cite{li2020learning}, train surrogate models with little robustness~\cite{zhang2024does}, or make the surrogate models more Bayesian~\cite{li2023making}.

\noindent\textbf{Generative modeling attacks} craft transferable adversarial examples by training generative models. Typical methods include Generative Adversarial Perturbations (GAP)~\cite{poursaeed2018generative}, Cross-Domain Transferable Perturbations (CDTP)~\cite{naseer2019cross}, and Attentive-Diversity Attack (ADA)~\cite{kim2022diverse}.

\noindent\textbf{Model ensemble attacks} draw from neural network training and generally combine the outputs of multiple surrogates to significantly enhance adversarial transferability. Typical methods include Stochastic Variance Reduction Ensemble attack (SVRE)~\cite{b14}, Adaptive Ensemble Attack (AdaEA)~\cite{chen2023adaptive}, and Common Weakness Attack (CWA)~\cite{b33}.

Each of the above attack types follows a specific perspective and has its own advantages.
In this study, we show that our AES method can be integrated with different types of attacks and their combinations to achieve improved transferability as well as stealthiness.

\subsection{Adversarial Defenses}

To mitigate the threat of adversarial attacks, numerous defense methods have been proposed. Among them, adversarial training (AT) methods~\cite{b8,b11,b40}, which iteratively incorporate adversarial examples into the training data, are widely used.
Although AT is effective even against white-box attacks, it suffers from high training costs. 
Existing research has commonly agreed that AT is not necessary against black-box attacks. 
Therefore, more efficient defenses have been explored against black-box attacks.
Specifically, Tram{\`e}r \textit{et al.}~\cite{b41} adopt AT but only generate one-time (transferable) adversarial examples for training.
Other defenses rely on input preprocessing.
Specifically, Liao \textit{et al.}~\cite{b45} utilize the High-Level Representation Guided Denoiser (HGD) to eliminate adversarial perturbations from input images. Xie \textit{et al.}~\cite{b46} employ random Resizing and Padding (R\&P) on adversarial examples to reduce their effectiveness. Cohen \textit{et al.}~\cite{b47} present the Randomized Smoothing (RS) method to obtain ImageNet classifiers with certified adversarial robustness. Moreover, Naseer \textit{et al.}~\cite{b48} leverage a Neural Representation Purifier (NRP) model to purify adversarial examples. Xu \textit{et al.}~\cite{b49} design Bit-Reduction (Bit-Red) and Spatial Smoothing as defenses against adversarial examples. 
In this study, we evaluate our AES against diverse target models with the above state-of-the-art defenses.

\input{tables/ASR_loss}

\section{Adversarial Example Soups (AES)}
\label{sec:method}

\subsection{Preliminary}

Let $x$ and $y$ be the benign input and its corresponding true label, respectively, while $\theta$ represents the parameters of the classifier $f$. $L(\theta,x,y)$ denotes the loss function of the classifier $f$, typically the cross-entropy loss function. The objective of the adversary is to find an adversarial example $x^{adv}$ that visually resembles but can mislead the model $f({{x}^{adv}};\theta )\ne y$ by maximizing $L(\theta,x,y)$. To maintain consistency with previous work~\cite{b21,b22}, we impose a constraint on the adversarial perturbation using the infinity norm, such that $\|x^{adv}-x\|_{\infty}\leq \varepsilon$. Therefore, the generation of adversarial examples can be transformed into the following constrained maximization problem:
\begin{equation}
\underset{x^{adv}}{\arg \max } L\left(\theta, x^{adv}, y\right), \quad \text{s.t.}\left\|x^{adv}-x\right\|_{\infty} \leq \varepsilon.
\label{eq1}
\end{equation}
Directly solving the optimization problem of Eq.~\ref{eq1} is quite complex. Therefore, Goodfellow \textit{et al.}~\cite{b8}, inspired by the training process of neural networks, propose the Fast Gradient Sign Method. This method increases the value of the loss function in the direction of gradient ascent to craft adversarial examples. 
Alexey \textit{et al.}~\cite{b50} extend the FGSM to an iterative version called I-FGSM, with the following update formula:

\begin{equation}
x_{0}^{adv}=x, x_{t+1}^{adv}=x_{t}^{a d v}+\alpha \cdot \operatorname{sign}\left(\nabla_{x} L\left(\theta, x_{t}^{adv}, y\right)\right),
\label{eq2}
\end{equation}
where $\alpha=\varepsilon/T$, in which $T$  is the number of iterations. FGSM and I-FGSM are simple to use and have good scalability, which are the basis of subsequent iterative attack methods. Based on Eq.~\ref{eq2}, the gradient stabilization attack is to optimize the gradient $\nabla_{x}L\left(\theta, x_{t}^{adv}, y\right)$ in Eq.~\ref{eq2}. The input transformation attack is to transform the input $x_{t}^{adv}$ at each iteration, and then compute the gradient of the transformed image. Feature destruction attacks use the features that the model focuses on to find a better gradient. Unlike existing transferable attacks, our approach focuses on the final output $x_{T}^{adv}$.

\subsection{Motivation of AES}
\label{sec:motivation}

The key assumption of our AES is that the multiple adversarial examples obtained in the optimization pipeline should reside in the same error basin~\cite{b27,b28,b29}.
Here we conduct exploratory experiments to show this assumption is satisfied in both hyperparameter tuning and stability testing, which are commonly conducted in existing work on transferable attacks.
Since the high dimensionality of images makes it difficult to directly analyze, we examine three later layers of the model that have lower dimensions, i.e., the internal features, loss values, and the final Attack Success Rates (ASR).
As shown in Table~\ref{table:loss}, in all these three metrics, we achieve very close results across different sessions of adversarial examples, suggesting that these adversarial examples are very likely to reside in the same error basin.
This finding is also consistent with that for ``model soups'', where multiple fine-tuned models initialized from the same pre-training often lie in the same error basin~\cite{b28}.

\begin{figure}[!t]
  \centering
    \includegraphics[width=1.0\linewidth]{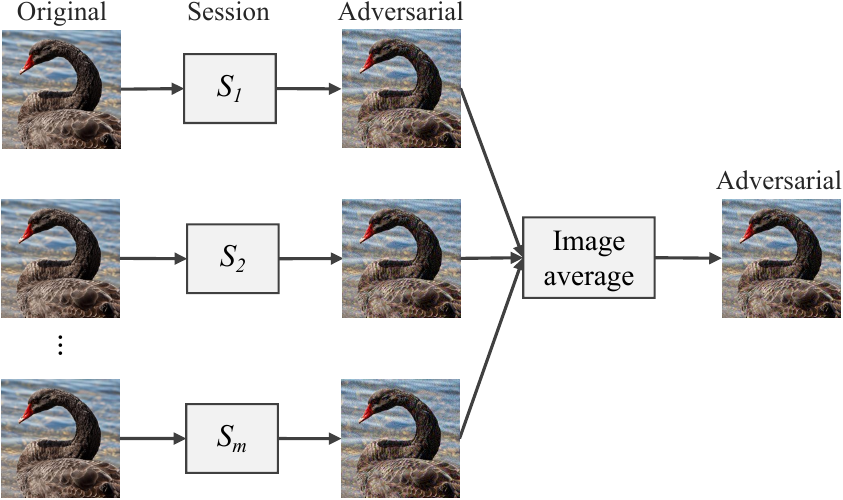}
    \caption{The framework of our Adversarial Example Soups (AES) attack. For each original image, AES averages its corresponding adversarial examples generated from $m$ different experimental sessions of hyperparameter tuning or stability testing. Each session takes a specific setting of hyperparameters.}
    \label{fig2}
\end{figure}

\subsection{Technical Details of AES}
As illustrated in Figure~\ref{fig2}, Adversarial Example Soups (AES) averages the adversarial examples from multiple sessions that correspond to different configurations of hyperparameters in hyperparameter tuning or repeated experiments in stability testing. 
Specifically, let $x_{i}^{adv}$ denote the adversarial example crafted from the session $S_{i}$, which refers to the adversarial example crafted under the hyperparameter setting $h_{i}$, where $i \in [1, m]$. For hyperparameter tuning, $h_{i}$ differs across sessions, with the specific settings shown in Table \ref{tab:hyperparameters}. For stability testing, $h_{i}$ remains the same across sessions and corresponds to the default settings of each attack method, with further details provided in Section \ref{sec:set}.
AES generates an adversarial example $x^{adv}$ following $x^{adv}=\sum\nolimits_{i=1}^{m}{w_{i}x_{i}^{adv}}$ with $\sum\nolimits_{i=1}^{m}{w_{i}}=1$, where $w_{i}$ represents the weight of the adversarial example from the session with $h_{i}$.
For the averaging strategy, we explore uniform, weighted, and greedy methods.
Specifically, the uniform method averages pixel values at corresponding positions of multiple adversarial images. 
The weighted and greedy methods further assign different weights to different sessions of adversarial examples based on their transferability ranking on a hold-out target model.
Adversarial examples by and find the greedy method yields a slightly better performance since it leverages more knowledge. 
Note that all three averaging strategies can effectively boost the transferability of baseline attacks, the uniform averaging way (where $w_{i}=1/m$) is adopted in the main experiments since it is the most simple and convenient (see more details in Section~\ref{sec:ablation}).

\noindent\textbf{AES-tune for hyperparameter tuning.} 
In order to select the hyperparameter for optimal attack transferability of a new attack method, it is common to conduct hyperparameter tuning. Specifically, in each session of hyperparameter tuning, a specific hyperparameter value is selected from a pre-defined range and used to generate an adversarial example.
Note that a hold-out target model is used for testing the transferability.
In this case, AES-tune averages adversarial examples from multiple sessions of hyperparameter tuning. 

\noindent\textbf{AES-rand for stability testing.}  
In order to make sure the transferability of a new attack is not sensitive to potential randomness in optimizing adversarial examples, stability testing is often conducted and results with variances are reported.
Such randomness generally comes from the non-deterministic GPU calculations and the change between dependent libraries, such as PyTorch and NumPy.
It can also come from the randomness of specific hyperparameters of an attack.
For example, DI implements stochastic transformation with probability $p$ at each iteration.
In this case, AES-rand averages adversarial examples from multiple sessions of stability testing.

In addition to the above AES-tune and AES-rand, there may be other forms of AES.
For example, the code for implementing different attack methods is publicly available, and adversarial images may be visualized anywhere.
In this case, we show that it is possible to average such in-the-wild adversarial examples that already yield comparable success to further boost their transferability (see detailed discussion in Section \ref{sec:AES-mix}).

\section{Experiments}
\label{sec:Exp}

In this section, we conduct extensive experiments to validate the effectiveness of our AES.
Specifically, we validate the global effectiveness of AES in improving the transferability of gradient stabilization, input transformation, and feature disruption attacks (Section~\ref{sec:trans}).
In addition, we demonstrate the effectiveness of AES in improving the stealthiness of different attacks in Section~\ref{sec:Steal}.
Finally, we conduct ablation studies to investigate the impact of important components and hyperparameters on the performance (Section~\ref{sec:ablation}).
In the following, we first describe the experimental setups. 

\subsection{Experimental Setups}
\label{sec:set}

\noindent\textbf{Dataset and models.} Following the common practice~\cite{b21,b34,zhao2020towards}, we randomly select 1000 images from the ImageNet validation set~\cite{b30} that are correctly classified.
For the surrogate model, we adopt four normally trained networks: Inception-v3 (Inc-v3)~\cite{b52}, Inception-v4 (Inc-v4)~\cite{b53}, Inception-Resnet-v2 (IncRes-v2)~\cite{b53}, and Resnet-v2-152 (Res-152)~\cite{b2}. For the target model, we consider those with diverse defenses or from different families, which are commonly believed to be more challenging than attacking normal CNN models. Specifically, we consider adversarially trained models~\cite{b41}, \textit{i.e.}, adv-Inception-v3 (Inc-v3$_{adv}$), ens3-adv-Inception-v3 (Inc-v3$_{ens3}$), ens4-adv-Inception-v3 (Inc-v3$_{ens4}$), and ens-adv-Inception-ResNet-v2 (IncRes-v2$_{ens}$), well-known defenses, \textit{i.e.},  HGD~\cite{b45}, R\&P~\cite{b46}, NIPS-r3\footnote{\url{https://github.com/anlthms/nips-2017/tree/master/mmd}}, NRP~\cite{b48}, and Bit-Red~\cite{b49}, as well as a Swin Transformer (Swin)~\cite{b54}.

\input{tables/hyperparameters}

\input{tables/gradient_based-sig}

\input{tables/input_transformation}

\input{tables/feature_disruption}
\noindent\textbf{Baseline attacks.} To demonstrate the global effectiveness of our AES, we select a variety of representative transferable attacks as our baselines. Specifically, for gradient stabilization attacks, we choose MI~\cite{b21}, NI~\cite{b22}, VMI~\cite{b34}, and PGN~\cite{b35}.
For input transformation attacks, we choose DIM~\cite{b36}, SIM~\cite{b22}, Admix~\cite{b37}, and SSA~\cite{b38}.
For feature disruption attacks. we choose FIA~\cite{b25} and NAA~\cite{b26}.

\noindent\textbf{Attack settings}. We follow the common practice~\cite{b22,b35,b36} to set the maximum magnitude of perturbation $\varepsilon=16$, the number of iterations $T=10$, and the step size $\alpha=1.6$. We adopt the default hyperparameter settings for different baseline attacks. Specifically, for MI and NI, we set the decay factor $\mu=1.0$. For VMI, we set the hyperparameter $\beta=1.5$ and the number of copies $N=20$. For PGN, we set the number of copies $N=20$, the balancing factor $\delta=0.5$ and the upper bound of neighborhood size $\zeta=3.0 \times \varepsilon$. For DIM, we set the transformation probability $p=0.5$. For SIM, we set the number of copies $N=5$. For Admix, we set the mix ratio $\eta=0.2$. For SSA, we set the tuning factor $\rho=0.5$ and the number of spectrum transformations $N=20$. For FIA, we set the drop probability $p_{d}=0.3$ and the ensemble number $N=30$. For NAA, we set the weighted attribution factor $\gamma=1.0$ and the integrated steps $I = 30$. 
For our AES, we set the number of sessions $m=10$, and the detailed hyperparameters for different attacks that form AES-tune are listed in Table~\ref{tab:hyperparameters}. All the above attacks have been described in Section~\ref{sec:adv-attack}.
Note that for other types of AES, the default settings of the baseline attacks are used.
For AES-rand, we only test randomness from the non-deterministic GPU calculations and the change between dependent libraries but not varied attack hyperparameters.

\subsection{AES Improves Transferability}
\label{sec:trans}
Here we evaluate the effectiveness of our AES in improving the transferability of three categories of attacks: gradient stability, input transformation, and feature disruption.
We consider the transferability settings with either a single surrogate model or an ensemble of surrogate models.

\input{tables/mean-std}
\input{tables/image_quality}

\begin{figure}[!t]
  \centering
    \includegraphics[width=1.0\columnwidth]{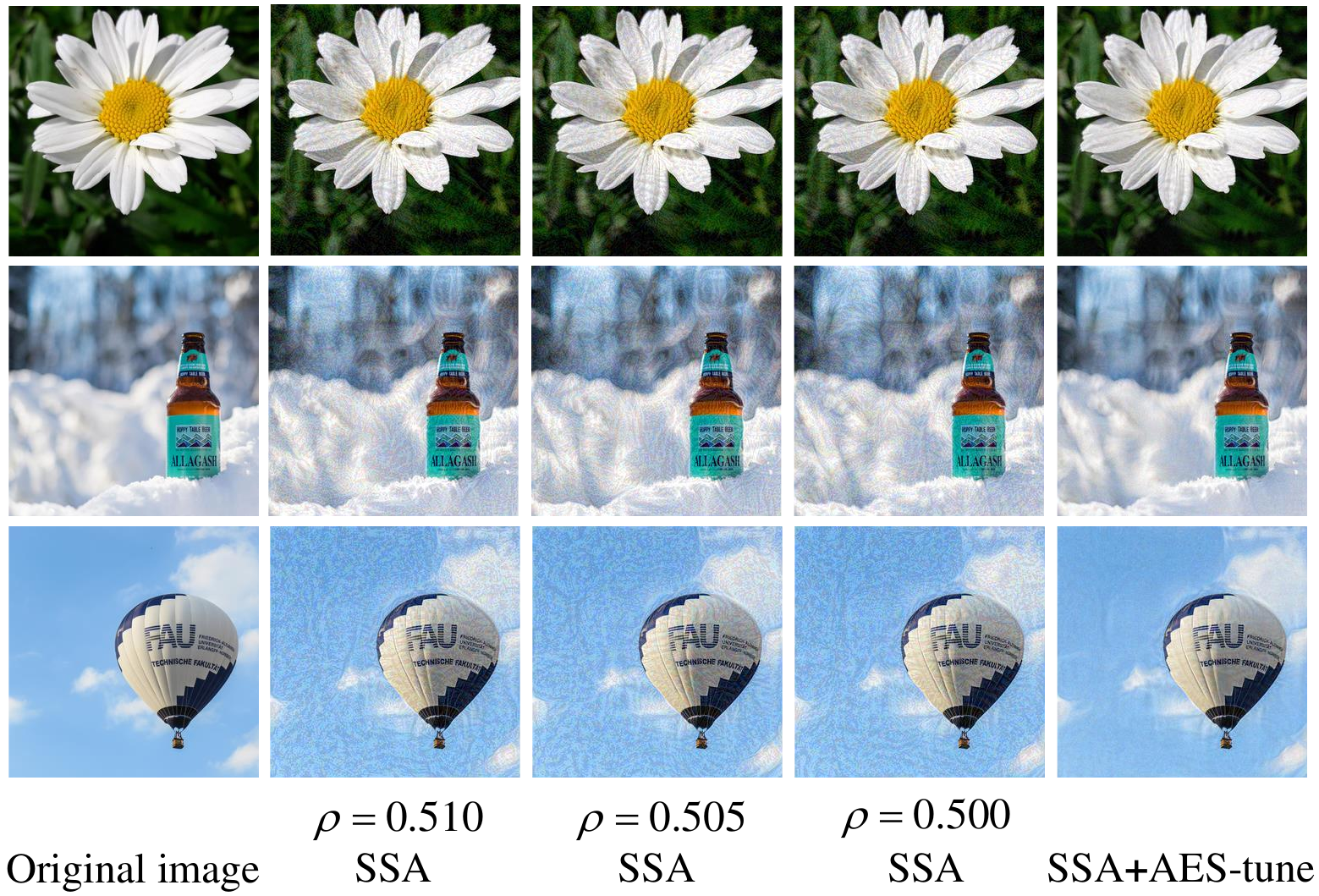}
    \caption{Visualizations of adversarial images generated by SSA with different hyperparameters vs. our AES. AES leads to higher image quality.}
   
    \label{fig:vis}
\end{figure}

Based on the results from Table~\ref{tab:gradient} to Table~\ref{tab:fea}, our AES improves all the 10 attacks (from three categories) on 10 diverse target models in both the settings of single surrogate and ensemble surrogate.
The improvement can be up to 13.3\%, for ensemble transfer with AES-tune for gradient stability attacks.
As expected, the results of ensemble transfer are generally better than those of the single transfer, but at the cost of more time and computations.
Note that for the ensemble surrogate transfer, we follow the common practice~\cite{b21} to combine the logit outputs of different models, here, Inc-v3, Inc-v4, InRes-v2, and Res-152.

In particular, we notice that the improvement on Swin is relatively small, suggesting that transferring to a different architecture is difficult.
This is consistent with the previous finding that changing the model architecture is normally more effective than applying a defense~\cite{b18}.
In almost all cases, AES-tune slightly outperforms AES-rand.
This may be because the change of the hyperparameter is normally larger than that in repeated experiments (with the same hyperparameter).

When comparing the three attack categories, we can observe that the improvement for gradient stability attacks under the single model setting is the smallest, \textit{i.e.}, 1.9\%-7.4\% for gradient stability vs. 5.2\%-12.3\% for input transformation vs. 5.6\%-10.8\% for feature disruption.  
This may be because the hyperparameter tuning for gradient stability attacks yields the least diverse results.
The results in Table~\ref{tab:mean} support our explanation, where gradient stability attacks yield the smallest fluctuation of transferability across different sessions of hyperparameters.
We confirm this finding also applies to single attack vs. ensemble attack and AES-rand vs. AES-tune.
In addition, we notice that in a few cases involving the PGN attack, AES even decreases the transferability, which is worth more in-depth exploration in the future.

\input{tables/Average_way}

\begin{figure}[!t]
	\centering
	\subfloat[AES-tune]{\includegraphics[width=\linewidth]{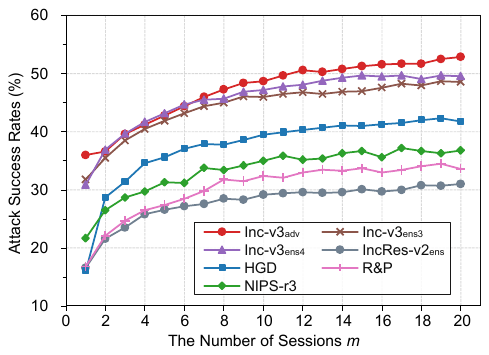} \label{fig4a}}
 
	\subfloat[AES-rand]
 {\includegraphics[width=\linewidth]{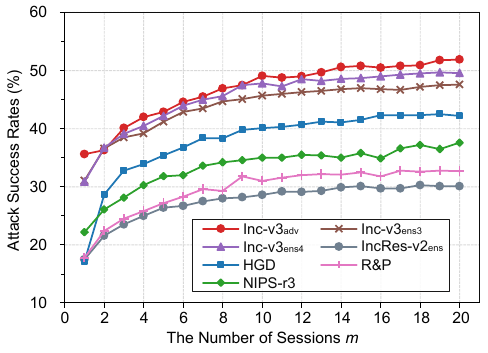}  \label{fig4b}}
	\caption{Ablation study on the number of sessions $m$. The surrogate model is Inc-v3.}
     \label{fig4}
\end{figure}

\subsection{AES Improves Stealthiness}
\label{sec:Steal}
 
Since AES takes the average of adversarial examples, it may cancel out specific pixel perturbations that are unstable across different sessions.
This suggests that AES may naturally lead to less noisy images, \textit{i.e.}, improving the attack stealthiness.
To validate this, we evaluate the visual quality of the adversarial images generated by AES using four popular metrics: Peak Signal-to-Noise Ratio (PSNR)~\cite{b18}, Structural Similarity Index Measure (SSIM)~\cite{b55}, Learned Perceptual Image Patch Similarity (LPIPS)~\cite{b56} and Frechet Inception Distance (FID)~\cite{b52}.
As shown in Table \ref{tab:imper}, both AES-tune and AES-rand substantially outperform the baseline attacks in all metrics. For example, in terms of the FID metric, DIM+AES-tune achieves the best FID score of 69.973, while DIM only achieves 95.472. 
The visualizations in Figure~\ref{fig:vis} further confirm that our AES leads to higher visual quality of the adversarial images.
In general, it is surprising that our AES can simultaneously improve transferability and stealthiness since most existing transferable attacks have to trade off these two properties~\cite{b18}.

\subsection{Ablation Studies}
\label{sec:ablation}

In this subsection, we conduct detailed ablation studies to explore the impact of hyperparameters of AES on its performance. 

\textbf{The number of sessions \textit{m}.} We use DIM as the baseline attack and generate adversarial examples on Inc-v3. The number of sessions $m$ ranges from 1 to 20. When $m = 1$, AES is equivalent to the naive DIM. As shown in Figure \ref{fig4a} and \ref{fig4b}, for both AES-tune and AES-rand, the transferability gradually increases as $m$ increases, and it becomes almost saturated after $m>10$.
It should be noted that a larger $m$ means a higher cost of time and computations.
In our main experiments, we set $m=10$ for a good trade-off between transferability and cost. 
In this case, the averaging operation on each image only costs 0.036s.

  \begin{figure}[!t]
  \centering
        \includegraphics[width=1.0\linewidth]{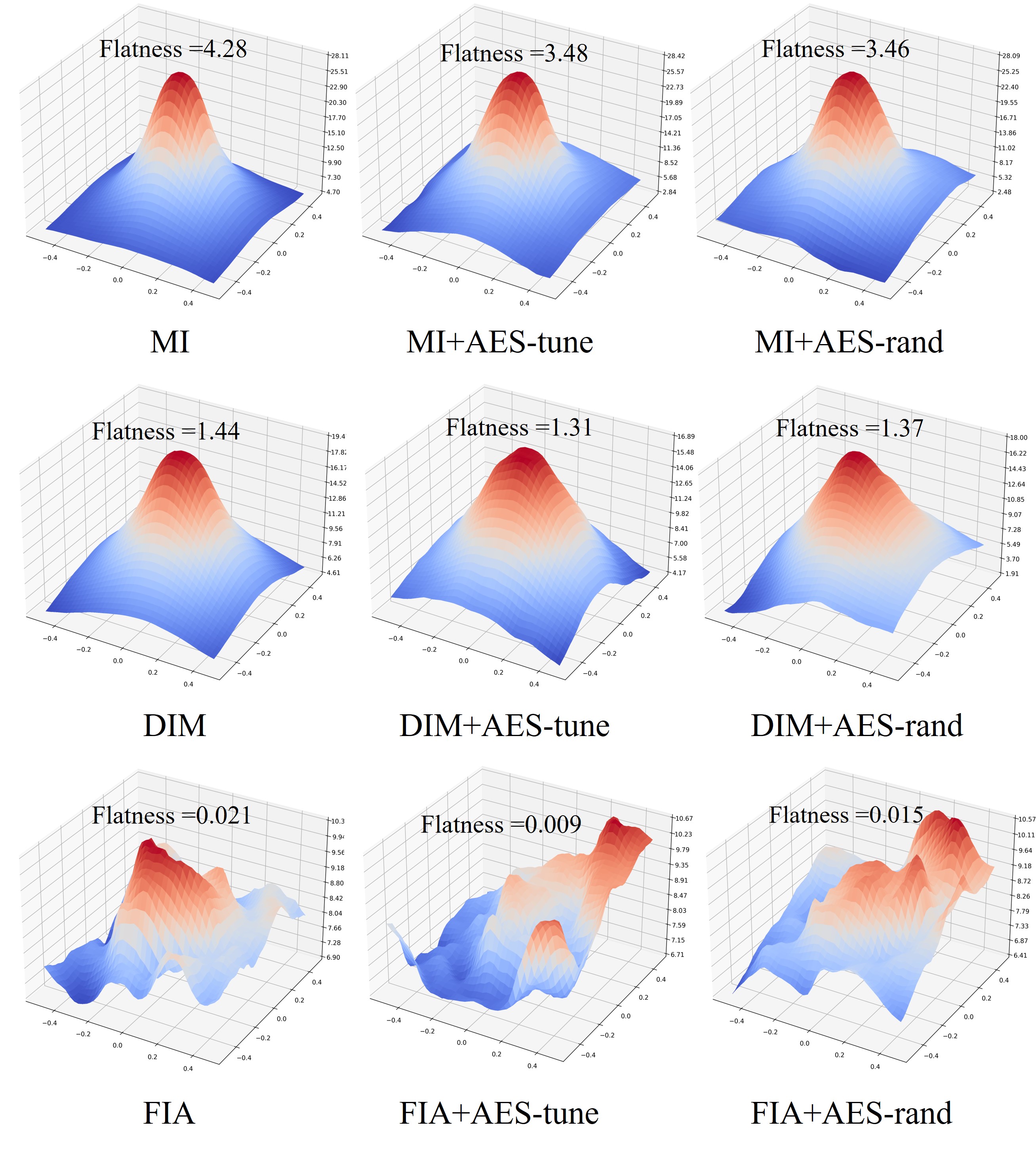}
    \caption{Visualization of loss surfaces along two random directions for one adversarial image and flatness (lower is better) averaged over 1000 adversarial images on Inc-v3. AES leads to flatter local maxima.}
    \label{fig:flatness}
\end{figure}

\begin{figure}[!t]
  \centering
    \includegraphics[width=1.0\columnwidth]{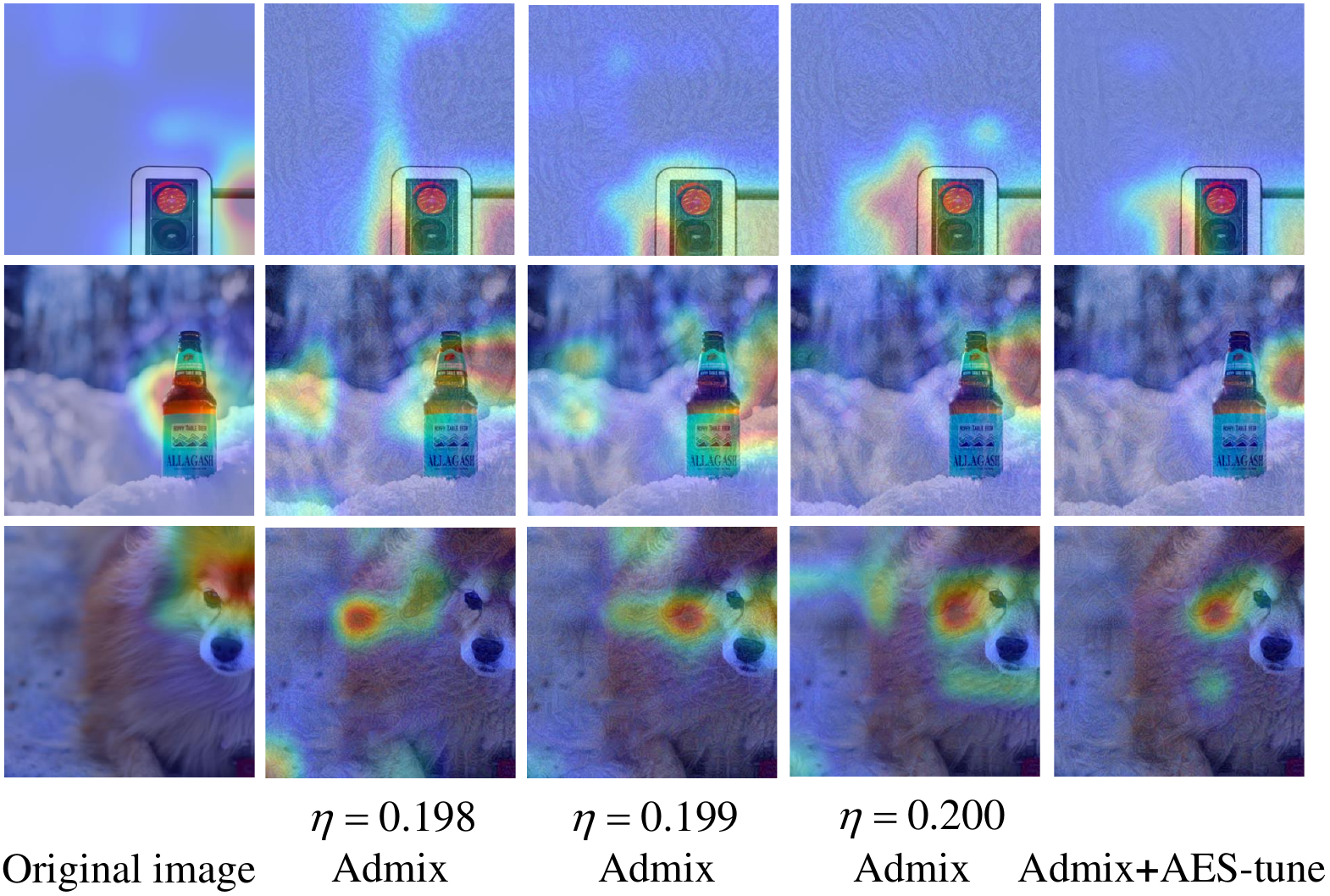}
    \caption{CAM attentions of adversarial images generated by Admix with different hyperparameters vs. AES. AES perturbs the central semantics, which can transfer across different models, and it cancels out unnecessary perturbations in the background.}
   
    \label{fig:cam}
\end{figure}

\textbf{Averaging strategy.} We consider three widely used averaging strategies: uniform, weighted, and greedy. 
Different from the uniform strategy in our main experiments, both the weighted and greedy strategies assign weights to adversarial examples in different sessions.
Here we use VMI since its hyperparameter, \textit{i.e.}, the number of copies $N$, is an integer.
Therefore, it is straightforward to determine the weights of different adversarial images according to $N$. 
Specifically, we vary $N$ from 16 to 25 to generate 10 sessions of adversarial examples on Inc-v3.
For the weighted strategy, the weights are determined by ranking the transferability of the 10 sessions of adversarial examples on Inc-v4, with the best-performed one assigned a weight of 25/205 and the worst assigned 16/205.
For the greedy strategy, only the top-5 adversarial examples according to the above ranking are used for averaging.
For a fair comparison and because the number of sessions has a significant impact on the performance, we still use 10 sessions, with the remaining 5 sessions randomly chosen from the original 10 sessions for $N=20$.
The results in Table \ref{tab:average} demonstrate that all three averaging strategies substantially improve transferability, with the greedy strategy achieving slightly better results than the other two.
Since the uniform averaging strategy does not require weighting allocation and sample selection, it is simpler and more efficient, so we mainly adopt this averaging method for AES. Naturally, we can also choose different averaging methods according to actual circumstances.

\input{tables/combined_method}

\section{Further Analysis and Discussion}
\label{sec:further}

\subsection{Explaining the Effectiveness of AES}
\label{sec:explain}

We first explain the effectiveness of AES from the perspective of loss flatness. Recent work~\cite{b29,cha2021swad} has demonstrated that averaging the parameters of models trained with different hyperparameter configurations can find flatter local minima, leading to improved generalizability of models. Similarly, research on adversarial examples also finds that the improved transferability can be explained by the flatness of loss landscape~\cite{springer2021little,zhang2024does}, and recent work has attempted to optimize transferability by forcing local flatness~\cite{b35,gubri2023going}.
Following this line of studies, we calculate the flatness of three representative attacks (i.e., MI, DI, and FIA) from the three attack categories.
Specifically, for each adversarial image, we draw its loss landscape by using it as the central point and move it along two random directions in the range of $[-0.5,0.5]$ with a step size of 0.025~\cite{b35}.
For each attack, the flatness is calculated by the difference in loss value between the sampling point and the center point, with a radius of 0.1 and a sampling number of 100~\cite{cha2021swad}. The results are then averaged over 1000 adversarial images.
As can be seen from Figure~\ref{fig:flatness}, for each attack, both our AES-tune and AES-rand achieve flatter local maxima than the baseline, explaining their better transferability as reported in Table~\ref{tab:gradient}-\ref{tab:fea}.

\input{tables/AES-mix}

\input{tables/image_quality_MNSA}

We further explain the effectiveness of AES from the perspective of model attention.
Specifically, we visualize the CAM~\cite{b57} attention map of a Res-50 classifier for clean vs. adversarial images, with their top-1 predicted label, \textit{i.e.}, correct labels for original images and incorrect labels for the adversarial images. 
As shown in Figure~\ref{fig:cam}, different hyperparameters of the Admix attack lead to attention maps that spread different local regions with a large overlap around the object. 
Specifically, after our AES is applied, the attention map concentrates more on the central object region.
This indicates that our AES perturbs more critical semantics but cancels out unstable perturbations that may not transfer across different models.
In addition, the fact that our AES cancels out unnecessary perturbations also explains its improved attack stealthiness regarding image quality.

\subsection{AES for Integrated Attacks}

In practice, different (categories of) transferable attacks can be integrated to achieve higher transferability. 
To shed light on the practical usefulness of AES, we evaluate it in such integrated scenarios, involving two categories of attacks: gradient stability and input transformation.
Since the integrated attacks contain multiple basic attacks, multiple hyperparameters can be tuned.
In our case, for AES-tune, we only tune the hyperparameter of one basic attack and keep the rest fixed.
Specifically, TI-DIM adjusts hyperparameter in DIM, SI-NI-DIM  in SIM, SSA-SI-DIM in SSA, and PGN-DIM in PGN. The specific details are shown in Table \ref{tab:hyperparameters}.
For AES-rand, the optimization is repeated multiple times as done in our main experiments.
As shown in Table \ref{tab:combine}, our AES achieves the highest average transferability in all cases.

\subsection{AES in the Wild}
\label{sec:AES-mix}

In the previous experiments, we focus on AES from the perspective of attack optimization, involving two common operations: hyperparameter tuning and stability testing.
There may also be other types of AES, as long as the underlying assumption of ``multiple adversarial examples lie in the same error basin'' is satisfied.
A promising type of AES in the real world is that we could leverage in-the-wild adversarial images to boost their transferability and stealthiness.
This is possible considering that the source code for implementing different attacks is publicly available, and different kinds of adversarial images may be visualized anywhere.

To validate the effectiveness of our AES in this scenario, we try directly averaging the adversarial examples from different attacks.
To this end, we select the adversarial examples from attacks that yield similar transferability on a hold-out target model.
Note that this selection can also be conducted at the sample level, by selecting the adversarial examples that yield similar loss values.
As shown in Table \ref{tab:mix} and \ref{tab:imper-MNSA}, AES for the pair of MI and NI or SIM and Admix consistently improves the transferability and stealthiness.

\subsection{Limitations and Broader Applications}
\label{sec:limitation}

In Section \ref{sec:explain}, we have followed the common practice to explain the effectiveness of our AES method from the perspective of loss flatness and model attention.
However, similar to most existing work on adversarial transferability, there still lacks a theoretical analysis. In the future, we would follow recent attempts~\cite{wang2020unified,zhang2024does} to conduct a more in-depth and theoretical study of our AES method.

Through our experiments, we have shown that our AES  outperforms existing methods against diverse defenses.
In the future, it would be promising to design specific defenses against AES. 
In general, we expect future stronger defenses can be used since our AES can be seen as a general post-processing operation of existing attacks. These defenses can also be combined to achieve a better performance, at the cost of more time and memories. Another promising way is to leverage a large set of AES adversarial examples during adversarial training~\cite{sitawarin2023defending}.

The paradigm of AES is not limited to images since it just averages the inputs from different sessions. Therefore, it is interesting for future work to extend AES to other data domains, such as text and speech. In particular, the discrete nature of text may introduce new challenges in ensuring the key assumption of AES that adversarial examples obtained in multiple sessions should reside in the same error basin.

\section{Conclusions}
\label{sec:conclu}

In this paper, we revisit the common recipe for optimizing transferability, which directly discards the adversarial examples obtained during the optimization process.
Specifically, we propose Adversarial Example Soups (AES), which reuse discarded adversarial examples for improved transferability and stealthiness.
We demonstrate the global effectiveness of AES in boosting 10 state-of-the-art transferable attacks and their combinations against 10 diverse (defensive) target models.
We also find that AES improves stealthiness since the perturbation variances are naturally reduced.
Beyond optimization, we discuss other types of AES, \textit{e.g.}, averaging multiple in-the-wild adversarial examples that yield comparable success to boost their transferability and stealthiness.

\section{Acknowledgement}
This research is supported by the National Key Research and Development Program of China (2023YFB3107400), the National Natural Science Foundation of China (62406240, 62376210, 62161160337, 62132011, U21B2018, U20A20177, U20B2049 and 62006181).

{
\bibliographystyle{IEEEtran}
\bibliography{reference}
}

\end{document}

%% file: tables/ASR_loss.tex
\begin{table}[!tbp]
\caption{The three measures across multiple sessions of adversarial examples obtained in hyperparameter tuning of $\mu$ in MI~\cite{b21} and stability testing yield small standard deviations. Inc-v3 is the surrogate model.}
\newcommand{\tabincell}[2]{\begin{tabular}{@{}#1@{}}#2\end{tabular}}

\renewcommand{\arraystretch}{1}
      \centering
      \resizebox{0.95\columnwidth}{!}{
        \begin{tabular}{ccccccc}
\toprule[1pt]
 \multirow{2}{*}{} & \multicolumn{3}{c}{Inc-v3$_{ens4}$} & \multicolumn{3}{c}{Inc-v2$_{ens}$}  \\ \cmidrule(lr){2-4} \cmidrule(lr){5-7}
 & Feature & Loss  & ASR & Feature & Loss  & ASR   \\ 
\toprule[1pt]

$\mu=0.91$                    & \textcolor{gray}{ 36.26} & \textcolor{gray}{ 0.95} & \textcolor{gray}{ 21.8} & \textcolor{gray}{ 24.99} & \textcolor{gray}{ 0.49} & \textcolor{gray}{ 10.9} \\
$\mu=0.92$                    & \textcolor{gray}{ 36.27} & \textcolor{gray}{ 0.96} & \textcolor{gray}{ 22.1} & \textcolor{gray}{ 24.96} & \textcolor{gray}{ 0.48} & \textcolor{gray}{ 10.7} \\
$\mu=0.93$                    & \textcolor{gray}{ 36.23} & \textcolor{gray}{ 0.92} & \textcolor{gray}{ 21.8} & \textcolor{gray}{ 24.94} & \textcolor{gray}{ 0.47} & \textcolor{gray}{ 11.0} \\
$\mu=0.94$                    & \textcolor{gray}{ 36.31} & \textcolor{gray}{ 0.94} & \textcolor{gray}{ 21.8} & \textcolor{gray}{ 24.95} & \textcolor{gray}{ 0.48} & \textcolor{gray}{ 11.4} \\
$\mu=0.95$                    & \textcolor{gray}{ 36.23} & \textcolor{gray}{ 0.93} & \textcolor{gray}{ 21.6} & \textcolor{gray}{ 24.95} & \textcolor{gray}{ 0.47} & \textcolor{gray}{ 10.3} \\
$\mu=0.96$                    & \textcolor{gray}{ 36.27} & \textcolor{gray}{ 0.95} & \textcolor{gray}{ 22.1} & \textcolor{gray}{ 24.95} & \textcolor{gray}{ 0.49} & \textcolor{gray}{ 11.0} \\
$\mu=0.97$                    & \textcolor{gray}{ 36.35} & \textcolor{gray}{ 0.94} & \textcolor{gray}{ 21.2} & \textcolor{gray}{ 24.94} & \textcolor{gray}{ 0.47} & \textcolor{gray}{ 10.8} \\
$\mu=0.98$                    & \textcolor{gray}{ 36.36} & \textcolor{gray}{ 0.93} & \textcolor{gray}{ 21.5} & \textcolor{gray}{ 24.95} & \textcolor{gray}{ 0.49} & \textcolor{gray}{ 10.8} \\
$\mu=0.99$                    & \textcolor{gray}{ 36.40} & \textcolor{gray}{ 0.94} & \textcolor{gray}{ 22.4} & \textcolor{gray}{ 24.97} & \textcolor{gray}{ 0.47} & \textcolor{gray}{ 10.6} \\
$\mu=1.00$                    & \textcolor{gray}{ 36.46} & \textcolor{gray}{ 0.95} & \textcolor{gray}{ 23.0} & \textcolor{gray}{ 24.93} & \textcolor{gray}{ 0.47} & \textcolor{gray}{ 11.4} \\   \hline
Average                          & 36.31                        & 0.94                        & 21.9                        & 24.95                        & 0.48                        & 10.9                        \\
Stand. dev.                              & 0.08                         & 0.01                        & 0.51                        & 0.02                         & 0.01                        & 0.34                        \\ \hline

                              & \textcolor{gray}{ 36.42} & \textcolor{gray}{ 0.96} & \textcolor{gray}{ 22.3} & \textcolor{gray}{ 24.95} & \textcolor{gray}{ 0.48} & \textcolor{gray}{ 11.2} \\
                              & \textcolor{gray}{ 36.42} & \textcolor{gray}{ 0.94} & \textcolor{gray}{ 22.3} & \textcolor{gray}{ 24.95} & \textcolor{gray}{ 0.47} & \textcolor{gray}{ 11.2} \\
                              & \textcolor{gray}{ 36.42} & \textcolor{gray}{ 0.93} & \textcolor{gray}{ 22.8} & \textcolor{gray}{ 24.98} & \textcolor{gray}{ 0.48} & \textcolor{gray}{ 11.2} \\
                              & \textcolor{gray}{ 36.40} & \textcolor{gray}{ 0.94} & \textcolor{gray}{ 22.0} & \textcolor{gray}{ 24.96} & \textcolor{gray}{ 0.48} & \textcolor{gray}{ 11.6} \\
                              & \textcolor{gray}{ 36.44} & \textcolor{gray}{ 0.93} & \textcolor{gray}{ 22.4} & \textcolor{gray}{ 24.94} & \textcolor{gray}{ 0.48} & \textcolor{gray}{ 11.1} \\
                              & \textcolor{gray}{ 36.50} & \textcolor{gray}{ 0.96} & \textcolor{gray}{ 21.9} & \textcolor{gray}{ 24.96} & \textcolor{gray}{ 0.48} & \textcolor{gray}{ 11.4} \\
                              & \textcolor{gray}{ 36.41} & \textcolor{gray}{ 0.94} & \textcolor{gray}{ 22.4} & \textcolor{gray}{ 24.94} & \textcolor{gray}{ 0.47} & \textcolor{gray}{ 11.2} \\
                              & \textcolor{gray}{ 36.43} & \textcolor{gray}{ 0.93} & \textcolor{gray}{ 21.8} & \textcolor{gray}{ 24.92} & \textcolor{gray}{ 0.47} & \textcolor{gray}{ 11.7} \\
                              & \textcolor{gray}{ 36.42} & \textcolor{gray}{ 0.95} & \textcolor{gray}{ 21.5} & \textcolor{gray}{ 24.95} & \textcolor{gray}{ 0.48} & \textcolor{gray}{ 11.3} \\
\multirow{-10}{*}{\tabincell{c}{Stability\\testing\\($\mu = 1.00$)}} & \textcolor{gray}{ 36.44} & \textcolor{gray}{ 0.91} & \textcolor{gray}{ 22.2} & \textcolor{gray}{ 24.99} & \textcolor{gray}{ 0.46} & \textcolor{gray}{ 10.8} \\ \hline
Average                           & 36.43                        & 0.94                        & 22.2                        & 24.95                        & 0.47                        & 11.3                        \\
Stand. dev.                           & 0.03                         & 0.02                        & 0.37                        & 0.02                         & 0.01                        & 0.25                      \\
\bottomrule[1pt]
\end{tabular}
}
\label{table:loss}
\end{table}

%% file: tables/hyperparameters.tex
\begin{table}[!t]
\newcommand{\tabincell}[2]{\begin{tabular}{@{}#1@{}}#2\end{tabular}}

\renewcommand{\arraystretch}{1}
      \centering
       \caption{The 10 hyperparameter configurations in AES-tune for different attacks.}
        \begin{tabular}{l|c}
\toprule[1pt]

Attack&Hyperparameter configurations\\

\midrule
MI~\cite{b21}&Decay factor $\mu=0.91, 0.92, \cdots, 1.00$\\
NI~\cite{b22} &Decay factor $\mu=0.91, 0.92, \cdots, 1.00$\\
VMI~\cite{b34} &Number of sampled examples $N=16, 17, \cdots, 25$\\
PGN~\cite{b35} &Balanced coefficient $\delta=0.491, 0.492, \cdots, 0.500$\\
\hline
DIM~\cite{b36} & Resize rate $r=1.132, 1.134, \cdots, 1.150$                   \\
SIM~\cite{b22}  & Decay factor $\mu=0.91, 0.92, \cdots, 1.00$      \\
Admix~\cite{b37}  & Mixing factor $\eta=0.191, 0.192, \cdots, 0.200$                \\
SSA~\cite{b38}  & Tuning factor $\rho=0.500, 0.505, \cdots, 0.545$                \\\hline
FIA~\cite{b25} & Drop probability $p_{d}=0.255, 0.260, \cdots, 0.300$               \\
NAA~\cite{b26}  &Weighted attribution factor $\gamma=0.91, 0.92, \cdots, 1.00$\\

\bottomrule[1pt]
\end{tabular}

\label{tab:hyperparameters}
\end{table}

%% file: tables/gradient_based-sig.tex
\begin{table*}[!t]
\small
\begin{center}
\caption{The success rates (\%) of gradient stabilization attacks with vs. without our AES.}
\resizebox{\textwidth}{!}{
\begin{tabular}{c|c|cccccccccc|c}
\hline
Surrogate & Attack & Inc-v3$_{adv}$ & Inc-v3$_{ens3}$ & Inc-v3$_{ens4}$ & IncRes-v2$_{ens}$ & HGD & R\&P & NIPS-r3 & NRP & Bit-Red & Swin& AVG\\
\hline\hline
\multirow{12}{*}{Inc-v3}    & MI         & 27.2          & 22.2          & 22.4         & 11.2          & 8.2           & 10.7          & 12.6          & 16.1          & 26.4          & 26.4          & 18.3          \\
                            & +AES-tune  & 31.7          & \textbf{26.8} & \textbf{28.5} & \textbf{15.2} & \textbf{16.1} & \textbf{14.2} & \textbf{15.5} & \textbf{17.7} & 26.5          & \textbf{27.8} & \textbf{22.0} \\
                            & +AES-rand  & \textbf{32.1} & 26.4          & 27.0          & 14.1          & 14.4          & 13.5          & \textbf{15.5} & 16.5          & \textbf{26.7} & 27.6          & 21.4          \\\cline{2-13}
                            & NI         & 27.8          & 23.1          & 23.1          & 11.7          & 8.3           & 11.3          & 13.6          & 16.8          & 26.6          & 30.2          & 19.3          \\
                            & +AES-tune  & 35.9          & 29.8          & \textbf{30.9} & \textbf{16.2} & \textbf{18.1} & \textbf{15.9} & \textbf{19.0} & 16.6          & \textbf{27.9} & 32.0          & \textbf{24.2} \\
                            & +AES-rand  & \textbf{36.0} & \textbf{30.0} & 29.7          & 15.6          & 17.0          & 14.7          & 18.3          & \textbf{16.8} & 27.9          & \textbf{32.9} & 23.9          \\\cline{2-13}
                            & VMI        & 45.1          & 42.5          & 42.1          & 25.4          & 23.8          & 23.7          & 30.7          & 23.5          & 34.5          & 43.7          & 33.5          \\
                            & +AES-tune & \textbf{51.0} & 45.1          & 46.1          & 29.2          & \textbf{33.1} & \textbf{29.1} & 32.3          & \textbf{26.1} & 35.3          & 44.7          & 37.2          \\
                            & +AES-rand & 50.8          & \textbf{46.1} & \textbf{46.3} & \textbf{30.0} & 33.0          & 28.4          & \textbf{32.5} & 25.6          & \textbf{35.5} & \textbf{45.2} & \textbf{37.3} \\\cline{2-13}
                            & PGN        & 71.4          & 65.4          & 64.5          & 45.5          & 38.5          & 46.0          & 53.2          & 44.8          & 52.9          & \textbf{62.3} & 54.5          \\
                            & +AES-tune & 75.5          & \textbf{71.2} & 74.1          & \textbf{56.8} & \textbf{62.9} & 54.3          & 60.1          & \textbf{48.3} & 56.2          & 59.3          & \textbf{61.9} \\
                            & +AES-rand & \textbf{76.1} & \textbf{71.2} & \textbf{74.7} & 56.5          & 62.4          & \textbf{54.5} & \textbf{60.3} & 48.0          & \textbf{56.6} & 58.2          & \textbf{61.9} \\\hline
\multirow{12}{*}{Inc-v4}    & MI         & 25.3          & 19.8          & 18.5          & 11.9          & 8.3           & 10.4          & 12.4          & 14.7          & 23.9          & 29.3          & 17.5          \\
                            & +AES-tune  & \textbf{27.9} & \textbf{23.5} & \textbf{24.0} & \textbf{13.3} & \textbf{14.7} & \textbf{14.1} & \textbf{13.7} & \textbf{15.5} & \textbf{25.8} & 30.3          & \textbf{20.3} \\
                            & +AES-rand  & 27.6          & \textbf{23.5} & 23.9          & 12.7          & 13.1          & 13.6          & 13.5          & 14.9          & 24.4          & \textbf{31.8} & 19.9          \\\cline{2-13}
                            & NI         & 24.4          & 19.6          & 18.9          & 11.1          & 7.7           & 10.2          & 12.2          & 14.5          & 25.2          & 30.9          & 17.5          \\
                            & +AES-tune  & \textbf{27.8} & \textbf{25.3} & \textbf{25.4} & \textbf{14.3} & \textbf{17.5} & \textbf{14.6} & \textbf{16.3} & 14.3          & 25.9          & 32.4          & \textbf{21.4} \\
                            & +AES-rand  & 26.0          & 23.7          & 23.7          & 12.9          & 15.1          & 12.9          & 14.9          & \textbf{15.2} & \textbf{26.4} & \textbf{33.9} & 20.5          \\\cline{2-13}
                            & VMI        & 40.7          & 41.7          & 40.4          & 27.2          & 26.0          & 27.7          & 31.2          & 23.4          & 34.1          & 49.1          & 34.2          \\
                            & +AES-tune & \textbf{45.5} & 45.8          & \textbf{46.7} & \textbf{32.6} & \textbf{35.9} & \textbf{33.5} & \textbf{36.3} & \textbf{26.1} & \textbf{35.9} & \textbf{51.0} & \textbf{38.9} \\
                            & +AES-rand & 45.1          & \textbf{45.9} & 46.4          & 31.9          & 34.8          & 32.4          & 35.5          & 25.9          & 35.5          & 50.1          & 38.4          \\\cline{2-13}
                            & PGN        & 65.6          & 67.1          & 64.3          & 49.0          & 38.7          & 48.1          & 55.3          & 46.6          & 54.2          & \textbf{70.9} & 56.0          \\
                            & +AES-tune & 69.8          & 71.1          & 70.5          & \textbf{60.4} & \textbf{63.6} & \textbf{56.9} & 60.3          & \textbf{51.1} & \textbf{55.8} & 64.2          & \textbf{62.4} \\
                            & +AES-rand & \textbf{71.2} & \textbf{71.5} & \textbf{71.5} & 59.6          & 62.9          & 56.7          & \textbf{60.6} & 50.1          & 55.2          & 64.4          & \textbf{62.4} \\\hline
\multirow{12}{*}{IncRes-v2} & MI         & 27.7          & 21.2          & 21.7          & 14.7          & 12.4          & 13.5          & 15.5          & 16.1          & 26.1          & 27.8          & 19.7          \\
                            & +AES-tune  & \textbf{31.2} & 26.3          & \textbf{27.5} & \textbf{21.1} & \textbf{22.2} & 18.1          & 19.8          & 16.0          & 25.7          & 29.2          & \textbf{23.7} \\
                            & +AES-rand  & 30.6          & \textbf{27.0} & 25.7          & 19.5          & 20.4          & \textbf{18.3} & \textbf{20.2} & \textbf{16.1} & \textbf{27.1} & \textbf{29.8} & 23.5          \\\cline{2-13}
                            & NI         & 27.2          & 18.7          & 18.4          & 12.6          & 10.5          & 11.8          & 13.9          & 14.7          & 24.5          & \textbf{28.2} & 18.1          \\
                            & +AES-tune  & \textbf{32.4} & \textbf{25.6} & \textbf{24.9} & \textbf{19.0} & \textbf{22.1} & \textbf{18.1} & \textbf{19.2} & \textbf{16.3} & 24.9          & 27.5          & \textbf{23.0} \\
                            & +AES-rand  & 30.2          & 24.1          & 23.8          & 17.1          & 19.7          & 15.9          & 17.3          & 15.9          & \textbf{25.5} & 27.4          & 21.7          \\\cline{2-13}
                            & VMI        & 47.2          & 48.7          & 44.5          & 37.5          & 36.0          & 34.9          & 37.5          & 23.8          & 35.4          & 47.4          & 39.3          \\
                            & +AES-tune & \textbf{53.4} & \textbf{54.7} & 50.5          & 46.7          & 47.8          & \textbf{42.6} & \textbf{44.0} & 26.9          & \textbf{37.2} & \textbf{48.5} & \textbf{45.2} \\
                            & +AES-rand & 53.3          & 53.7          & \textbf{50.7} & \textbf{47.4} & \textbf{47.9} & 42.3          & 43.9          & \textbf{27.0} & 37.1          & 48.1          & 45.1          \\\cline{2-13}
                            & PGN        & 75.1          & 75.3          & 70.8          & 66.3          & 58.2          & 63.6          & 66.9          & 51.5          & 58.5          & \textbf{69.8} & 65.6          \\
                            & +AES-tune & 77.7          & 78.4          & 75.6          & \textbf{74.7} & \textbf{73.8} & 69.6          & 72.0          & \textbf{56.2} & \textbf{61.7} & 64.8          & \textbf{70.5} \\
                            & +AES-rand & \textbf{79.4} & \textbf{78.5} & \textbf{75.7} & 73.9          & \textbf{73.8} & \textbf{69.7} & \textbf{72.3} & 56.0          & 61.6          & 64.3          & \textbf{70.5} \\\hline
\multirow{12}{*}{Res-152}   & MI         & 28.9          & 27.0          & 25.9          & 15.4          & 17.7          & 15.4          & 18.8          & 19.2          & 27.0          & 31.0          & 22.6          \\
                            & +AES-tune  & 32.5          & 30.1          & 29.9          & 19.7          & 25.5          & 19.1          & 20.8          & 19.1          & 26.4          & \textbf{32.0} & 25.5          \\
                            & +AES-rand  & \textbf{32.8} & \textbf{30.9} & \textbf{31.1} & \textbf{19.8} & \textbf{26.1} & \textbf{19.4} & \textbf{20.8} & \textbf{19.4} & \textbf{28.0} & 31.8          & \textbf{26.0} \\\cline{2-13}
                            & NI         & 31.3          & 28.1          & 25.5          & 16.0          & 16.2          & 16.2          & 19.1          & 19.0          & 26.5          & 32.4          & 23.0          \\
                            & +AES-tune  & 36.8          & \textbf{34.9} & \textbf{35.1} & \textbf{21.9} & \textbf{32.4} & \textbf{23.3} & \textbf{26.6} & \textbf{19.7} & 28.6          & 35.4          & \textbf{29.5} \\
                            & +AES-rand  & \textbf{37.2} & 32.9          & 33.7          & 20.5          & 29.4          & 21.8          & 24.8          & 19.2          & \textbf{29.3} & \textbf{36.4} & 28.5          \\\cline{2-13}
                            & VMI        & 45.4          & 46.6          & 43.1          & 32.5          & 38.4          & 32.3          & 37.3          & 27.2          & 35.7          & 47.2          & 38.6          \\
                            & +AES-tune & \textbf{50.6} & \textbf{51.1} & 47.6          & \textbf{37.3} & \textbf{44.6} & \textbf{36.9} & \textbf{42.0} & 29.1          & 36.3          & \textbf{48.5} & \textbf{42.4} \\
                            & +AES-rand & 49.9          & 50.6          & \textbf{47.7} & 36.0          & 44.0          & 36.2          & 41.3          & \textbf{29.4} & \textbf{36.4} & 48.1          & 42.0          \\\cline{2-13}
                            & PGN        & 71.3          & 69.0          & 68.1          & 58.4          & 61.8          & 58.1          & 63.3          & 49.0          & 54.3          & \textbf{62.2} & 61.6          \\
                            & +AES-tune & 72.7          & \textbf{72.3} & 69.8          & \textbf{63.2} & \textbf{67.2} & 59.8          & \textbf{64.1} & \textbf{52.6} & \textbf{57.9} & 57.4          & \textbf{63.7} \\
                            & +AES-rand & \textbf{73.2} & 71.6          & \textbf{70.2} & 61.7          & 66.5          & \textbf{60.4} & 63.7          & 52.4          & 57.8          & 57.6          & 63.5     \\    
\hline
\multirow{12}{*}{Ensemble} & MI         & 42.4          & 43.0          & 40.9          & 27.5          & 33.3          & 26.8          & 33.3          & 23.5          & 34.1          & 54.5          & 35.9          \\
                           & +AES-tune  & \textbf{52.5} & \textbf{53.3} & \textbf{50.9} & \textbf{36.1} & \textbf{51.4} & \textbf{36.0} & \textbf{41.0} & \textbf{24.7} & 34.2          & 58.8          & \textbf{43.9} \\
                           & +AES-rand  & 51.2          & 52.5          & 49.9          & 35.3          & 47.6          & 34.9          & 39.5          & \textbf{24.7} & \textbf{34.6} & \textbf{59.4} & 43.0          \\\cline{2-13}
                           & NI         & 44.5          & 44.8          & 41.0          & 26.6          & 29.1          & 27.3          & 32.3          & 24.4          & 34.2          & 57.4          & 36.2          \\
                           & +AES-tune  & \textbf{62.9} & \textbf{60.2} & \textbf{57.4} & \textbf{42.7} & \textbf{57.1} & \textbf{42.3} & \textbf{47.7} & 25.1          & 34.8          & \textbf{64.5} & \textbf{49.5} \\
                           & +AES-rand  & 58.8          & 56.0          & 53.1          & 38.6          & 48.8          & 37.4          & 43.1          & \textbf{25.4} & \textbf{36.2} & 63.2          & 46.1          \\\cline{2-13}
                           & VMI        & 67.4          & 69.8          & 67.9          & 56.0          & 62.0          & 55.8          & 60.0          & 40.3          & 49.5          & 74.2          & 60.3          \\
                           & +AES-tune & \textbf{73.4} & \textbf{73.9} & \textbf{73.4} & \textbf{62.0} & \textbf{69.6} & 62.4          & 64.0          & 40.7          & 50.0          & \textbf{74.9} & \textbf{64.4} \\
                           & +AES-rand & 72.2          & 73.3          & 73.1          & 61.7          & 69.3          & \textbf{62.9} & \textbf{65.0} & \textbf{41.2} & \textbf{50.8} & 74.8          & \textbf{64.4} \\\cline{2-13}
                           & PGN        & 88.8          & 89.2          & 88.1          & 82.7          & 84.0          & 83.1          & 85.6          & 74.2          & 78.7          & \textbf{89.6} & 84.4          \\
                           & +AES-tune & 90.9          & 90.3          & \textbf{89.8} & \textbf{86.4} & 89.2          & \textbf{85.9} & \textbf{87.6} & 78.5          & \textbf{79.8} & 87.9          & \textbf{86.6} \\
                           & +AES-rand & \textbf{91.0} & \textbf{90.5} & 89.4          & 86.3          & \textbf{89.3} & \textbf{85.9} & 87.0          & \textbf{78.7} & 79.1          & 88.0          & 86.5          \\
                           \hline

\end{tabular}}
\label{tab:gradient}
\end{center}
\end{table*}

%% file: tables/input_transformation.tex
\begin{table*}[!t]
\small
\begin{center}
\caption{The success rates (\%) of input transformation attacks with vs. without our AES.}
\resizebox{\textwidth}{!}{
\begin{tabular}{c|c|cccccccccc|c}
\hline
Surrogate & Attack & Inc-v3$_{adv}$ & Inc-v3$_{ens3}$ & Inc-v3$_{ens4}$ & IncRes-v2$_{ens}$ & HGD & R\&P & NIPS-r3 & NRP & Bit-Red & Swin &AVG\\
\hline\hline
\multirow{12}{*}{Inc-v3}    & DIM          & 36.0          & 31.8          & 30.9          & 16.6          & 16.0          & 16.6          & 22.0          & 18.5          & 30.7          & 38.3          & 25.7          \\
                            & +AES-tune   & \textbf{49.3} & \textbf{45.8} & 47.2          & \textbf{28.8} & 39.1          & 31.0          & \textbf{35.5} & \textbf{21.8} & 32.3          & \textbf{42.1} & \textbf{37.3} \\
                            & +AES-rand   & 47.4          & \textbf{45.8} & \textbf{47.4} & 28.7          & \textbf{39.4} & \textbf{31.1} & 34.3          & 20.8          & \textbf{32.6} & 41.8          & 36.9          \\\cline{2-13}
                            & SIM          & 46.4          & 39.4          & 38.6          & 22.9          & 20.1          & 21.4          & 27.5          & 25.6          & 37.8          & 41.3          & 32.1          \\
                            & +AES-tune   & \textbf{56.2} & \textbf{47.5} & \textbf{47.9} & \textbf{29.6} & \textbf{35.8} & \textbf{28.1} & \textbf{36.0} & \textbf{27.1} & 38.3          & \textbf{42.2} & \textbf{38.9} \\
                            & +AES-rand   & 54.5          & 45.6          & 45.8          & 28.1          & 31.7          & 26.9          & 33.0          & 26.5          & \textbf{38.6} & 41.8          & 37.3          \\\cline{2-13}
                            & Admix        & 50.2          & 45.3          & 45.4          & 27.0          & 24.9          & 25.4          & 33.0          & 30.3          & 40.9          & 46.3          & 36.9          \\
                            & +AES-tune & \textbf{62.8} & 55.5          & 56.4          & 37.9          & 47.5          & 37.8          & 42.4          & 31.3          & \textbf{44.2} & 47.4          & 46.3          \\
                            & +AES-rand & 62.0          & \textbf{57.4} & \textbf{57.9} & \textbf{38.5} & \textbf{49.3} & \textbf{38.3} & \textbf{42.8} & \textbf{31.8} & \textbf{44.2} & \textbf{48.6} & \textbf{47.1} \\\cline{2-13}
                            & SSA          & 64.3          & 56.6          & 55.8          & 35.4          & 32.9          & 36.3          & 42.6          & 34.3          & 45.4          & 56.0          & 46.0          \\
                            & +AES-tune   & \textbf{71.7} & \textbf{67.4} & \textbf{67.7} & \textbf{49.9} & \textbf{59.9} & \textbf{48.6} & \textbf{54.8} & \textbf{40.1} & \textbf{48.9} & 56.1          & \textbf{56.5} \\
                            & +AES-rand   & 71.6          & \textbf{67.4} & 67.4          & 49.9          & 59.7          & 48.5          & 54.4          & 39.4          & 47.9          & \textbf{56.4} & 56.3          \\\hline
\multirow{12}{*}{Inc-v4}    & DIM          & 28.7          & 26.1          & 25.8          & 15.8          & 16.3          & 16.8          & 19.2          & 16.5          & 27.0          & 38.6          & 23.1          \\
                            & +AES-tune   & \textbf{38.3} & \textbf{39.9} & 38.5          & \textbf{27.7} & \textbf{37.2} & \textbf{29.0} & 28.8          & \textbf{18.4} & 28.9          & 44.8          & \textbf{33.2} \\
                            & +AES-rand   & 37.1          & 38.6          & \textbf{38.8} & 26.4          & 36.7          & 28.4          & \textbf{30.0} & 17.6          & \textbf{30.0} & \textbf{45.0} & 32.9          \\\cline{2-13}
                            & SIM          & 45.8          & 47.7          & 42.8          & 28.8          & 26.4          & 27.7          & 33.6          & 24.1          & 37.9          & 51.2          & 36.6          \\
                            & +AES-tune   & \textbf{55.3} & \textbf{56.6} & \textbf{53.3} & \textbf{37.6} & \textbf{44.0} & \textbf{38.4} & \textbf{43.1} & 25.9          & 40.2          & 54.2          & \textbf{44.9} \\
                            & +AES-rand   & 53.6          & 55.6          & 52.3          & 35.9          & 37.8          & 36.3          & 39.5          & \textbf{26.9} & \textbf{40.3} & \textbf{55.7} & 43.4          \\\cline{2-13}
                            & Admix        & 50.8          & 53.5          & 49.5          & 32.6          & 33.7          & 33.0          & 39.6          & 29.5          & 41.4          & 56.9          & 42.1          \\
                            & +AES-tune & 59.4          & \textbf{64.0} & \textbf{61.7} & \textbf{46.8} & \textbf{56.7} & \textbf{47.2} & \textbf{52.2} & \textbf{31.8} & 43.5          & \textbf{58.8} & \textbf{52.2} \\
                            & +AES-rand & \textbf{60.8} & 62.2          & 60.6          & 45.2          & 55.2          & 45.8          & 51.6          & 31.4          & \textbf{43.7} & 56.2          & 51.3          \\\cline{2-13}
                            & SSA          & 59.5          & 57.1          & 55.6          & 36.0          & 30.8          & 37.9          & 44.7          & 35.1          & 44.6          & \textbf{64.2} & 46.6          \\
                            & +AES-tune   & \textbf{66.2} & \textbf{66.8} & 65.4          & 50.0          & 58.9          & 51.5          & 54.6          & 41.0          & \textbf{48.8} & 61.3          & 56.5          \\
                            & +AES-rand   & 65.6          & 66.7          & \textbf{66.1} & \textbf{51.4} & \textbf{59.6} & \textbf{52.2} & \textbf{56.5} & \textbf{41.2} & 47.5          & 61.1          & \textbf{56.8} \\\hline
\multirow{12}{*}{IncRes-v2} & DIM          & 34.6          & 33.7          & 31.7          & 22.1          & 21.2          & 22.2          & 23.6          & 18.8          & 28.7          & 38.2          & 27.5          \\
                            & +AES-tune   & \textbf{41.0} & \textbf{42.0} & 38.9          & \textbf{37.0} & \textbf{39.2} & \textbf{35.8} & 36.6          & \textbf{19.2} & \textbf{29.2} & \textbf{38.6} & \textbf{35.8} \\
                            & +AES-rand   & 40.8          & 41.8          & \textbf{39.5} & \textbf{37.0} & 38.9          & 35.2          & \textbf{37.0} & 19.0          & \textbf{29.2} & 38.4          & 35.7          \\\cline{2-13}
                            & SIM          & 57.0          & 56.5          & 47.9          & 39.4          & 38.1          & 37.5          & 41.6          & 29.9          & 42.3          & 46.4          & 43.7          \\
                            & +AES-tune   & \textbf{67.4} & \textbf{64.7} & \textbf{58.0} & \textbf{54.1} & \textbf{59.9} & \textbf{50.0} & \textbf{53.3} & \textbf{32.0} & \textbf{44.5} & \textbf{49.6} & \textbf{53.4} \\
                            & +AES-rand   & 65.5          & 63.5          & 55.6          & 51.0          & 53.3          & 46.2          & 50.0          & 31.9          & 43.2          & 48.1          & 50.8          \\\cline{2-13}
                            & Admix        & 66.5          & 68.9          & 62.3          & 52.7          & 52.9          & 49.7          & 54.7          & 37.1          & 50.8          & 57.4          & 55.3          \\
                            & +AES-tune & \textbf{77.2} & \textbf{75.4} & \textbf{71.9} & \textbf{68.9} & \textbf{73.3} & \textbf{65.2} & \textbf{68.2} & \textbf{40.4} & \textbf{51.4} & \textbf{59.9} & \textbf{65.2} \\
                            & +AES-rand & 75.3          & \textbf{75.4} & 69.4          & 66.9          & 70.7          & 63.2          & 66.4          & 39.7          & 50.4          & 57.7          & 63.5          \\\cline{2-13}
                            & SSA          & 67.2          & 67.6          & 61.7          & 55.1          & 56.3          & 54.2          & 58.0          & 38.8          & 49.5          & \textbf{63.7} & 57.2          \\
                            & +AES-tune   & 73.6          & 73.1          & \textbf{69.6} & \textbf{68.1} & 69.2          & \textbf{63.7} & 66.0          & 43.3          & \textbf{53.6} & 61.9          & 64.2          \\
                            & +AES-rand   & \textbf{73.9} & \textbf{73.4} & 69.3          & 67.4          & \textbf{70.1} & 63.6          & \textbf{66.2} & \textbf{44.9} & 53.4          & 60.9          & \textbf{64.3} \\\hline
\multirow{12}{*}{Res-152}   & DIM          & 41.0          & 42.4          & 40.0          & 25.8          & 34.4          & 26.4          & 31.3          & 23.8          & 34.3          & 45.4          & 34.5          \\
                            & +AES-tune   & \textbf{53.6} & \textbf{56.6} & \textbf{52.8} & \textbf{42.7} & \textbf{54.3} & \textbf{45.4} & \textbf{48.4} & \textbf{25.4} & \textbf{38.3} & \textbf{50.4} & \textbf{46.8} \\
                            & +AES-rand   & 52.8          & 54.9          & 51.9          & \textbf{42.7} & 53.9          & 45.0          & 47.1          & 25.2          & 37.0          & 50.1          & 46.1          \\\cline{2-13}
                            & SIM          & 47.7          & 45.6          & 43.3          & 29.3          & 33.0          & 29.6          & 34.2          & 28.5          & 37.4          & 44.9          & 37.4          \\
                            & +AES-tune   & \textbf{56.9} & \textbf{55.1} & \textbf{51.5} & \textbf{38.7} & \textbf{49.8} & \textbf{39.5} & \textbf{44.6} & 28.3          & 39.5          & \textbf{46.3} & \textbf{45.0} \\
                            & +AES-rand   & 56.8          & 52.8          & 50.5          & 36.1          & 44.5          & 36.3          & 42.0          & \textbf{29.2} & \textbf{39.7} & 45.9          & 43.4          \\\cline{2-13}
                            & Admix        & 45.9          & 47.2          & 43.3          & 31.3          & 35.9          & 30.8          & 36.2          & 30.5          & 39.1          & 45.1          & 38.5          \\
                            & +AES-tune & \textbf{56.9} & \textbf{56.5} & \textbf{54.3} & \textbf{44.3} & \textbf{53.9} & \textbf{42.1} & \textbf{49.1} & \textbf{31.9} & \textbf{41.7} & \textbf{49.2} & \textbf{48.0} \\
                            & +AES-rand & 55.0          & 54.6          & 52.7          & 43.3          & 52.8          & 40.6          & 46.9          & 31.2          & 40.3          & 48.5          & 46.6          \\\cline{2-13}
                            & SSA          & 66.2          & 63.0          & 60.2          & 45.5          & 54.7          & 47.6          & 54.2          & 41.2          & 49.2          & \textbf{59.6} & 54.1          \\
                            & +AES-tune   & \textbf{70.2} & 68.1          & \textbf{66.1} & 58.1          & \textbf{67.5} & \textbf{57.6} & \textbf{61.8} & \textbf{45.9} & \textbf{52.5} & 57.4          & \textbf{60.5} \\
                            & +AES-rand   & 69.5          & \textbf{68.4} & 65.9          & \textbf{58.4} & 67.3          & 56.9          & 60.7          & 45.1          & 52.2          & 58.7          & 60.3          \\
\hline
\multirow{12}{*}{Ensemble} & DIM          & 63.7          & 72.2          & 67.1          & 53.0          & 63.0          & 57.1          & 62.7          & 36.1          & 49.0          & 74.5          & 59.8          \\
                           & +AES-tune   & \textbf{81.9} & 82.6          & \textbf{80.0} & 73.6          & 81.6          & 76.4          & 76.9          & 39.7          & 53.4          & \textbf{78.3} & 72.4          \\ 
                           & +AES-rand   & 80.8          & \textbf{82.9} & 79.4          & \textbf{74.3} & \textbf{81.7} & \textbf{76.6} & \textbf{77.4} & \textbf{40.6} & \textbf{55.3} & 78.1          & \textbf{72.7} \\\cline{2-13}
                           & SIM          & 76.5          & 79.1          & 76.0          & 58.7          & 65.9          & 59.9          & 66.4          & 46.5          & 57.8          & 79.6          & 66.6          \\
                           & +AES-tune   & \textbf{87.4} & \textbf{87.0} & \textbf{85.3} & \textbf{73.4} & \textbf{85.9} & \textbf{75.4} & \textbf{80.6} & 47.6          & \textbf{61.0} & \textbf{83.2} & \textbf{76.7} \\
                           & +AES-rand   & 84.9          & 86.5          & 83.4          & 68.5          & 80.3          & 69.9          & 76.6          & \textbf{47.7} & 60.4          & 82.6          & 74.1          \\\cline{2-13}
                           & Admix        & 74.5          & 80.3          & 78.1          & 64.1          & 74.1          & 63.9          & 69.6          & 46.9          & 58.1          & 80.2          & 69.0          \\
                           & +AES-tune & \textbf{84.0} & \textbf{86.0} & \textbf{84.4} & \textbf{75.4} & \textbf{84.5} & \textbf{75.2} & \textbf{78.8} & \textbf{49.8} & \textbf{61.1} & \textbf{82.1} & \textbf{76.1} \\
                           & +AES-rand & 80.2          & 83.8          & 81.8          & 70.5          & 81.3          & 70.7          & 75.3          & 47.2          & 59.5          & 81.6          & 73.2          \\\cline{2-13}
                           & SSA          & 89.3          & 88.9          & 85.8          & 77.1          & 81.3          & 80.1          & 83.7          & 62.6          & 72.4          & \textbf{89.9} & 81.1          \\
                           & +AES-tune   & 91.4          & 90.7          & 89.4          & 85.6          & 90.7          & 86.7          & 88.4          & \textbf{72.2} & 75.6          & 88.0          & 85.9          \\
                           & +AES-rand   & \textbf{91.6} & \textbf{91.0} & \textbf{89.5} & \textbf{85.9} & \textbf{91.2} & \textbf{86.8} & \textbf{88.7} & 71.2          & \textbf{76.0} & 88.8          & \textbf{86.1} \\
\hline
\end{tabular}}
\label{tab:input}
\end{center}
\end{table*}

%% file: tables/feature_disruption.tex
\begin{table*}[!t]
\small
\begin{center}
\caption{The success rates (\%) of feature disruption attacks with vs. without our AES.}
\resizebox{\textwidth}{!}{
\begin{tabular}{c|c|cccccccccc|c}
\hline
Surrogate & Attack & Inc-v3$_{adv}$ & Inc-v3$_{ens3}$ & Inc-v3$_{ens4}$ & IncRes-v2$_{ens}$ & HGD & R\&P & NIPS-r3 & NRP & Bit-Red & Swin &AVG\\
\hline\hline
\multirow{6}{*}{Inc-v3}    & FIA        & 53.3          & 35.8          & 35.5          & 20.9          & 10.5          & 19.0          & 25.6          & 24.4          & 40.9          & 46.3          & 31.2          \\
                           & +AES-tune & \textbf{59.8} & \textbf{47.3} & \textbf{46.8} & \textbf{30.7} & \textbf{33.0} & \textbf{29.6} & \textbf{36.3} & \textbf{26.2} & 44.4          & \textbf{47.3} & \textbf{40.1} \\
                           & +AES-rand & 59.0          & 45.0          & 44.7          & 29.2          & 31.3          & 28.7          & 35.0          & 25.0          & \textbf{44.5} & 46.6          & 38.9          \\\cline{2-13}
                           & NAA        & 63.7          & 52.5          & 50.6          & 32.6          & 20.4          & 33.1          & 40.0          & 33.5          & 47.1          & 58.2          & 43.2          \\
                           & +AES-tune & 69.5          & 60.4          & 60.8          & 45.3          & 49.3          & 44.7          & 50.1          & 37.5          & 49.9          & \textbf{58.3} & 52.6          \\
                           & +AES-rand & \textbf{69.6} & \textbf{61.0} & \textbf{60.8} & \textbf{46.8} & \textbf{49.9} & \textbf{46.2} & \textbf{50.2} & \textbf{38.4} & \textbf{49.9} & \textbf{58.3} & \textbf{53.1} \\\hline
\multirow{6}{*}{Inc-v4}    & FIA        & 45.1          & 38.4          & 36.7          & 20.5          & 16.7          & 21.2          & 27.4          & 23.4          & 39.1          & \textbf{54.7} & 32.3          \\
                           & +AES-tune & \textbf{55.3} & \textbf{49.6} & \textbf{47.7} & \textbf{29.9} & \textbf{37.1} & \textbf{31.9} & \textbf{37.8} & 24.9          & \textbf{42.2} & 54.3          & \textbf{41.1} \\
                           & +AES-rand & 54.4          & 49.1          & 46.7          & 29.2          & 36.9          & 30.4          & 36.6          & \textbf{25.5} & 41.1          & 54.4          & 40.4          \\\cline{2-13}
                           & NAA        & 53.3          & 51.8          & 47.3          & 32.2          & 28.8          & 31.6          & 37.5          & 31.2          & 43.7          & 63.5          & 42.1          \\
                           & +AES-tune & \textbf{61.0} & \textbf{59.0} & \textbf{58.0} & \textbf{43.9} & 51.1          & \textbf{44.3} & \textbf{49.2} & 32.0          & 44.2          & 63.3          & \textbf{50.6} \\
                           & +AES-rand & 60.8          & 58.3          & 57.5          & 43.3          & \textbf{51.4} & 43.9          & 48.5          & \textbf{32.4} & \textbf{44.5} & \textbf{64.1} & 50.5          \\\hline
\multirow{6}{*}{IncRes-v2} & FIA        & \textbf{63.5}          & 50.6          & 45.0          & 34.7          & 23.4          & 31.2          & 40.5          & 29.5          & 50.8          & \textbf{47.4} & 41.7          \\
                           & +AES-tune & 62.6 & \textbf{56.0} & \textbf{53.1} & \textbf{45.4} & \textbf{48.2} & \textbf{41.3} & \textbf{47.9} & \textbf{31.7} & \textbf{51.4} & 46.8          & \textbf{48.4} \\
                           & +AES-rand & 62.0          & 55.0          & 53.0          & 45.1          & 46.9          & 40.3          & 46.7          & 30.7          & 50.4          & 46.7          & 47.7          \\\cline{2-13}
                           & NAA        & 64.4          & 59.3          & 55.9          & 51.0          & 38.0          & 46.8          & 51.0          & 40.2          & 50.9          & \textbf{57.4} & 51.5          \\
                           & +AES-tune & \textbf{65.8}          & 64.0          & 60.5          & 58.6          & 57.4          & \textbf{55.6} & 56.5          & \textbf{43.2} & \textbf{53.9} & 55.7          & 57.1          \\
                           & +AES-rand & \textbf{65.8} & \textbf{64.2} & \textbf{61.0} & \textbf{59.3} & \textbf{59.5} & 55.5          & \textbf{57.2} & 42.8          & 52.0          & 56.1          & \textbf{57.3} \\\hline
\multirow{6}{*}{Res-152}   & FIA        & 57.9          & 48.9          & 43.8          & 28.4          & 33.8          & 30.8          & 36.8          & 28.9          & 46.9          & 50.5          & 40.7          \\
                           & +AES-tune & 65.6          & 58.8          & \textbf{56.9} & \textbf{44.7} & \textbf{54.0} & \textbf{43.3} & \textbf{50.2} & 30.8          & 47.5          & \textbf{52.0} & \textbf{50.4} \\
                           & +AES-rand & \textbf{65.7} & \textbf{59.1} & 56.1          & 43.6          & 53.6          & 42.6          & 48.5          & \textbf{31.0} & \textbf{48.2} & 50.6          & 49.9          \\\cline{2-13}
                           & NAA        & 65.6          & 61.4          & 57.6          & 47.1          & 49.9          & 48.3          & 53.1          & 39.2          & 49.8          & 63.1          & 53.5          \\
                           & +AES-tune & 71.8          & 68.5          & \textbf{66.0} & \textbf{58.0} & 66.6          & 57.8          & 62.3          & \textbf{42.7} & 52.7          & 63.0          & 60.9          \\
                           & +AES-rand & \textbf{72.0} & \textbf{69.0} & 66.1          & 57.7          & \textbf{67.8} & \textbf{58.7} & \textbf{62.7} & 42.6          & \textbf{53.0} & \textbf{63.4} & \textbf{61.3} \\ \hline
\multirow{6}{*}{Ensemble} & FIA       & 72.0          & 62.4          & 57.6          & 41.0          & 38.1          & 39.4          & 48.5          & 36.3          & 57.1          & 53.9          & 50.6          \\
                          & +AES-tune & \textbf{77.8} & \textbf{70.2} & \textbf{68.3} & 55.9          & \textbf{67.6} & \textbf{56.7} & 61.8          & 39.8          & \textbf{58.9} & \textbf{56.6} & \textbf{61.4} \\
                          & +AES-rand & 77.4          & 70.1          & 68.1          & \textbf{56.5} & 66.7          & 56.0          & \textbf{62.1} & \textbf{40.8} & 58.6          & 55.4          & 61.2          \\\cline{2-13}
                          & NAA       & 75.9          & 71.3          & 70.5          & 57.9          & 57.5          & 57.2          & 63.0          & 47.2          & 59.5          & 65.3          & 62.5          \\
                          & +AES-tune & 81.9          & 79.3          & \textbf{77.9} & \textbf{70.4} & 76.6          & 70.3          & 72.4          & 51.6          & \textbf{63.4} & 65.4          & 70.9          \\
                          & +AES-rand & \textbf{82.1} & \textbf{79.5} & 77.7          & 70.3          & \textbf{77.5} & \textbf{70.6} & \textbf{74.0} & \textbf{52.2} & 63.0          & \textbf{66.6} & \textbf{71.4} \\

\hline
\end{tabular}}

\label{tab:fea}
\end{center}
\end{table*}

%% file: tables/mean-std.tex
\begin{table}[tbp]
\caption{Diversity of success rates for different attacks. Gradient stability attacks yield the lowest diversity, represented by the lowest standard deviations.
AES-rand on Inc-v3 is used.}
\newcommand{\tabincell}[2]{\begin{tabular}{@{}#1@{}}#2\end{tabular}}
\renewcommand{\arraystretch}{1}
      \centering
      \resizebox{\columnwidth}{!}{
        \begin{tabular}{c|ccccc}
\toprule[1pt]
Attack & Inc-v3$_{adv}$ & Inc-v3$_{ens3}$ & Inc-v3$_{ens4}$ & IncRes-v2$_{ens}$  \\
\midrule
MI    & \textcolor{gray}{27.1}(±0.62)          & \textcolor{gray}{22.6}(±0.58)          & \textcolor{gray}{22.2}(±0.37)          & \textcolor{gray}{11.3}(±0.25)          \\
NI    & \textcolor{gray}{28.3}(±0.66)          & \textcolor{gray}{22.5}(±0.44)          & \textcolor{gray}{22.5}(±0.57)          & \textcolor{gray}{11.4}(±0.42)          \\
VMI   & \textcolor{gray}{45.1}(±0.37)          & \textcolor{gray}{41.5}(±0.47)          & \textcolor{gray}{41.3}(±0.41)          & \textcolor{gray}{24.9}(±0.42)          \\
PGN   & \textcolor{gray}{70.5}(±0.67)          & \textcolor{gray}{64.9}(±0.58)          & \textcolor{gray}{65.7}(±0.68)          & \textcolor{gray}{45.2}(±0.64)          \\
DIM   & \textcolor{gray}{35.4}(±0.59)          & \textcolor{gray}{31.7}(±\textbf{1.18}) & \textcolor{gray}{31.4}(±\textbf{0.86}) & \textcolor{gray}{17.0}(±0.66)          \\
SIM   & \textcolor{gray}{46.5}(±0.59)          & \textcolor{gray}{39.3}(±0.52)          & \textcolor{gray}{38.2}(±0.40)          & \textcolor{gray}{23.1}(±0.51)          \\
Admix & \textcolor{gray}{52.1}(±0.43)          & \textcolor{gray}{45.9}(±0.72)          & \textcolor{gray}{45.2}(±0.69)          & \textcolor{gray}{26.9}(±0.45)          \\
SSA   & \textcolor{gray}{63.0}(±\textbf{0.69}) & \textcolor{gray}{56.4}(±0.50)          & \textcolor{gray}{56.3}(±0.67)          & \textcolor{gray}{35.2}(±0.74)          \\
FIA   & \textcolor{gray}{54.2}(±0.65)          & \textcolor{gray}{35.8}(±0.50)          & \textcolor{gray}{35.4}(±0.50)          & \textcolor{gray}{19.8}(±0.69)          \\
NAA   & \textcolor{gray}{63.5}(±0.51)          & \textcolor{gray}{52.0}(±0.83)          & \textcolor{gray}{51.7}(±0.58)          & \textcolor{gray}{32.5}(±\textbf{0.78})  \\

\bottomrule[1pt]
\end{tabular}
}
\label{tab:mean}
\end{table}

%% file: tables/image_quality.tex
\begin{table}[!t]
\caption{The stealthiness of different attacks with or without our AES. The surrogate model is Inc-v3.}
\newcommand{\tabincell}[2]{\begin{tabular}{@{}#1@{}}#2\end{tabular}}

\renewcommand{\arraystretch}{1}
      \centering
        \begin{tabular}{c|cccccc}
\toprule[1pt]
Attack  &PSNR$\uparrow  $&SSIM$\uparrow$   &LPIPS$\downarrow$   &FID$\downarrow$\\
\midrule

MI           & 26.954          & 0.6691          & 0.1348          & 83.858           \\
+AES-tune    & \textbf{28.643} & \textbf{0.7441} & \textbf{0.1038} & \textbf{77.102}  \\
+AES-rand    & 27.892          & 0.7219          & 0.1154          & 78.732           \\\hline
NI           & 26.921          & 0.6653          & 0.1385          & 98.459           \\
+AES-tune    & \textbf{29.261} & \textbf{0.7594} & \textbf{0.0985} & \textbf{84.759}  \\
+AES-rand    & 28.360          & 0.7235          & 0.1121          & 88.527           \\\hline
VMI          & 27.090          & 0.6840          & 0.1262          & 96.902           \\
+AES-tune   & 27.627          & 0.7103          & 0.1166          & 93.024           \\
+AES-rand   & \textbf{27.655} & \textbf{0.7125} & \textbf{0.1151} & \textbf{92.519}  \\\hline
PGN          & 26.630          & 0.6730          & 0.1272          & 115.034          \\
+AES-tune   & 29.934          & 0.8489          & \textbf{0.0659} & 101.437          \\
+AES-rand   & \textbf{29.935} & \textbf{0.8490} & 0.0660          & \textbf{99.778}  \\\hline
DIM          & 26.920          & 0.6686          & 0.1330          & 95.472           \\
+AES-tune   & 31.944          & 0.8501          & 0.0605          & 69.973           \\
+AES-rand   & \textbf{32.161} & \textbf{0.8566} & \textbf{0.0581} & \textbf{67.438}  \\\hline
SIM          & 26.879          & 0.6688          & 0.1357          & 112.549          \\
+AES-tune   & \textbf{28.649} & \textbf{0.7507} & \textbf{0.1042} & \textbf{104.025} \\
+AES-rand   & 27.952          & 0.7218          & 0.1156          & 106.001          \\\hline
Admix        & 26.825          & 0.6697          & 0.1336          & 115.356          \\
+AES-tune & 29.624          & 0.8077          & 0.0814          & 99.651           \\
+AES-rand & \textbf{29.722} & \textbf{0.8123} & \textbf{0.0801} & \textbf{97.342}  \\\hline
SSA          & 26.701          & 0.6628          & 0.1298          & 110.153          \\
+AES-tune   & \textbf{30.380} & \textbf{0.8507} & \textbf{0.0639} & 93.475           \\
+AES-rand   & 30.323          & 0.8493          & 0.0642          & \textbf{91.640}  \\\hline
FIA          & 26.560          & 0.6692          & 0.1369          & 136.777          \\
+AES-tune   & \textbf{28.748} & 0.7711          & \textbf{0.1016} & 126.076          \\
+AES-rand   & 28.747          & \textbf{0.7713} & 0.1017          & \textbf{125.552} \\\hline
NAA          & 26.545          & 0.6698          & 0.1369          & 123.369          \\
+AES-tune   & 28.990          & 0.7849          & \textbf{0.0941} & 111.337          \\
+AES-rand   & \textbf{29.001} & \textbf{0.7853} & 0.0943          & \textbf{108.836} \\

\bottomrule[1pt]
\end{tabular}
\label{tab:imper}
\end{table}

%% file: tables/Average_way.tex
\begin{table}[tbp]
\caption{Ablation study on the three averaging strategies: uniform, weighted, and greedy. These three strategies average the 10 sessions with different $N$.}
\newcommand{\tabincell}[2]{\begin{tabular}{@{}#1@{}}#2\end{tabular}}
\renewcommand{\arraystretch}{1}
      \centering
      \resizebox{\columnwidth}{!}{
        \begin{tabular}{c|cccc|c}
\toprule[1pt]
VMI & Inc-v3$_{adv}$ & Inc-v3$_{ens3}$ & Inc-v3$_{ens4}$ & IncRes-v2$_{ens}$  & AVG\\
\midrule

N=16                & 44.7          & 41.8          & 40.9          & 24.5          & 38.0          \\
N=17                & 45.5 & 40.3          & 40.6          & 24.6          & 37.8          \\
N=18                & 44.5          & 41.2          & 40.8          & 23.5          & 37.5          \\
N=19                & 44.9          & 41.3          & 41.8          & 24.6          & 38.2          \\
N=20                & 45.2          & 41.9          & 40.2          & 25.0          & 38.1          \\
N=21                & 45.2          & 41.1          & 42.5 & 24.7          & 38.4          \\
N=22                & 45.3          & 42.3 & 41.7          & 25.2          & 38.6          \\
N=23                & 45.3          & 41.3          & 41.3          & 25.0          & 38.2          \\
N=24                & 44.3          & 41.4          & 41.2          & 25.2          & 38.0          \\
N=25                & 44.5          & 41.4          & 41.1          & 25.6 & 38.2          \\ \hline
Uniform     & 51.0          & 45.1          & 46.1          & 29.2        & 42.9          \\  
Weighted    & 51.1          & 45.1          & 45.8          & 29.2          & 42.8          \\ 
Greedy      & \textbf{51.9}         & \textbf{46.3}         & \textbf{47.9}         & \textbf{30.0}        & \textbf{44.0}          \\

\bottomrule[1pt]
\end{tabular}
}
\label{tab:average}
\end{table}

%% file: tables/combined_method.tex
\begin{table*}[tb]
\small
\begin{center}
\caption{The success rates (\%) of AES for integrated attacks. The surrogate model is Inc-v3.}
\resizebox{\textwidth}{!}{
\begin{tabular}{c|cccccccccc|c}
\hline
 Attack & Inc-v3$_{adv}$ & Inc-v3$_{ens3}$ & Inc-v3$_{ens4}$ & IncRes-v2$_{ens}$ & HGD & R\&P & NIPS-r3 & NRP & Bit-Red & Swin & AVG\\
\hline\hline
TI-DIM     & 45.9          & 48.2          & 46.7          & 32.8          & 39.5          & 34.2          & 38.2          & 25.7          & 38.5          & 39.9          & 39.0          \\
+AES-tune   & 56.8          & 54.5          & \textbf{54.5} & 37.4          & 42.1          & \textbf{41.4} & \textbf{44.2} & \textbf{27.6} & \textbf{40.0} & 39.5          & 43.8          \\
+AES-rand   & \textbf{56.9} & \textbf{54.9} & 53.6          & \textbf{39.7} & \textbf{43.9} & 40.0          & 43.4          & 26.7          & 38.6          & \textbf{41.9} & \textbf{44.0} \\ \hline
SI-NI-DIM  & 52.3          & 47.0          & 45.0          & 28.7          & 26.8          & 27.6          & 36.7          & 28.0          & 42.1          & \textbf{49.2} & 38.3          \\
+AES-tune   & 58.1          & 56.8          & \textbf{57.1} & 40.8          & 48.0          & 40.6          & 45.5          & 30.7          & 42.2          & 45.5          & 46.5          \\
+AES-rand   & \textbf{60.1} & \textbf{57.5} & 56.7          & \textbf{41.3} & \textbf{49.5} & \textbf{41.3} & \textbf{46.4} & \textbf{32.5} & \textbf{42.7} & 47.0          & \textbf{47.5} \\ \hline
SSA-SI-DIM & 85.4          & 83.7          & 81.9          & 66.0          & 72.1          & 70.3          & 75.4          & 55.2          & 67.2          & 72.4          & 73.0          \\
+AES-tune   & \textbf{88.8} & \textbf{86.9} & \textbf{85.4} & \textbf{74.8} & 80.2          & 76.3          & \textbf{80.3} & 59.8          & 68.4          & 72.0          & 77.3          \\
+AES-rand   & 88.7          & 86.8          & 85.2          & 74.3          & \textbf{80.6} & \textbf{77.8} & \textbf{80.3} & \textbf{60.2} & \textbf{70.2} & \textbf{72.7} & \textbf{77.7} \\ \hline
PGN-DIM    & 77.7          & 75.8          & 75.3          & 58.4          & 56.2          & 61.3          & 66.9          & 53.5          & 63.0          & \textbf{74.2} & 66.2          \\
+AES-tune   & \textbf{82.4} & \textbf{80.4} & 80.3          & 68.0          & \textbf{75.2} & 66.9          & \textbf{72.8} & 58.0          & \textbf{64.9} & 68.2          & \textbf{71.7} \\
+AES-rand   & 81.6          & \textbf{80.4} & \textbf{80.8} & \textbf{68.2} & 74.9          & \textbf{67.4} & 71.5          & \textbf{59.0} & 64.2          & 68.4          & 71.6         \\
\hline
\end{tabular}}

\label{tab:combine}
\end{center}
\end{table*}

%% file: tables/AES-mix.tex
\begin{table*}[tb]
\small
\begin{center}
\caption{The success rates (\%) of AES in the wild that directly averages adversarial images from two attacks. The surrogate model is an ensemble of Inc-v3, Inc-v4, IncRes-v2, and Res-152.}
\resizebox{\textwidth}{!}{
\begin{tabular}{c|cccccccccc|c}
\hline
 Attack & Inc-v3$_{adv}$ & Inc-v3$_{ens3}$ & Inc-v3$_{ens4}$ & IncRes-v2$_{ens}$ & HGD & R\&P & NIPS-r3 & NRP & Bit-Red & Swin & AVG\\
\hline\hline
MI       & 42.6          & 42.8          & 40.8          & 26.4          & 33.3          & 26.8          & 33.3          & 22.7          & 33.9          & 56.0          & 35.9          \\
NI       & 45.5          & 45.5          & 39.6          & 27.3          & 29.1          & 27.3          & 32.3          & 23.1          & \textbf{35.1} & 57.7          & 36.3          \\
AES-MI\&NI & \textbf{49.6} & \textbf{53.3} & \textbf{49.2} & \textbf{37.8} & \textbf{50.4} & \textbf{35.6} & \textbf{40.2} & \textbf{26.6} & 35.0          & \textbf{61.8} & \textbf{44.0} \\ \hline
SIM      & 76.2          & 79.2          & 75.3          & 58.3          & 65.9          & 59.9          & 66.4          & 45.7          & 58.0          & 79.9          & 66.5          \\
Admix    & 75.2          & 81.3          & 77.7          & 64.0          & 74.1          & 63.9          & 69.6          & 46.0          & 57.8          & 79.8          & 68.9          \\
AES-SIM\&Admix  & \textbf{79.3} & \textbf{84.7} & \textbf{81.7} & \textbf{70.5} & \textbf{82.0} & \textbf{70.7} & \textbf{72.5} & \textbf{50.9} & \textbf{60.9} & \textbf{81.5} & \textbf{73.5} \\
\hline
\end{tabular}}

\label{tab:mix}
\end{center}
\end{table*}

%% file: tables/image_quality_MNSA.tex
\begin{table}[!t]
\caption{The stealthiness of AES in the wild that directly averages adversarial images from two attacks. The surrogate model is an ensemble of Inc-v3, Inc-v4, IncRes-v2, and Res-152.}
\newcommand{\tabincell}[2]{\begin{tabular}{@{}#1@{}}#2\end{tabular}}

\renewcommand{\arraystretch}{1}
      \centering
      \resizebox{0.9\columnwidth}{!}{
        \begin{tabular}{c|cccccc}
\toprule[1pt]
Attack  &PSNR$\uparrow  $&SSIM$\uparrow$   &LPIPS$\downarrow$   &FID$\downarrow$\\
\midrule

MI             & 27.025          & 0.6717          & 0.1360          & 85.810          \\
NI             & 26.922          & 0.6648          & 0.1425          & 88.191          \\
AES-MI\&NI     & \textbf{28.406} & \textbf{0.7262} & \textbf{0.1110} & \textbf{79.572} \\ \hline
SIM            & 26.851          & 0.6672          & 0.1349          & 108.315         \\
Admix          & 26.905          & 0.6734          & 0.1299          & 104.881         \\
AES-SIM\&Admix & \textbf{28.338} & \textbf{0.7345} & \textbf{0.1054} & \textbf{97.248} \\

\bottomrule[1pt]
\end{tabular}
}
\label{tab:imper-MNSA}
\end{table}